\documentclass[11pt,american,aip,floatfix,jcp,reprint,longbibliography]{revtex4-2}

\usepackage[utf8]{inputenc}
\usepackage{graphicx}
\usepackage{dcolumn}
\usepackage{bm}
\usepackage{hyperref}
\usepackage[final]{pdfpages}
\usepackage{pgffor}

\makeatletter
\AtBeginDocument{\let\LS@rot\@undefined}
\makeatother

\usepackage{amsmath}
\usepackage{amsfonts}
\usepackage{mathtools}
\usepackage{xcolor}
\usepackage[linesnumbered,ruled,vlined]{algorithm2e}

\SetCommentSty{mycommfont}

\SetKwInput{KwInput}{Input}                
\SetKwInput{KwOutput}{Output}

\makeatletter
\let\@fnsymbol\@fnsymbol@latex
\@booleanfalse\altaffilletter@sw
\makeatother

\newcommand{\Real}{\mathbb{R}}
\newcommand{\Int}{\mathbb{Z}_+}
\newcommand{\hyp}{h_\theta}
\newcommand{\expected}[2]{\underset{#1}{\mathbb{E}}\left[#2\right]}
\newcommand{\Loss}{\mathcal{L}}
\newcommand{\LossAdv}{\mathcal{L}_\mathrm{adv}}
\newcommand{\Mean}[3]{\frac{1}{#1} \displaystyle{\sum_{#2}^{#3}}}
\newcommand{\Var}[1]{\sigma^2_{#1}}

\DeclarePairedDelimiter{\norm}{\lVert}{\rVert}

\begin{document}

\title{Differentiable sampling of molecular geometries with uncertainty-based adversarial attacks}

\author{Daniel Schwalbe-Koda}
\thanks{D.S.-K. and A.R.T. contributed equally to this work}
\author{Aik Rui Tan}
\thanks{D.S.-K. and A.R.T. contributed equally to this work}
\author{Rafael G\'omez-Bombarelli}
\email{rafagb@mit.edu}

\affiliation{%
 Department of Materials Science and Engineering, Massachusetts Institute of Technology, Cambridge, MA 02139
}%

\date{\today}

\begin{abstract}
    Neural network (NN) interatomic potentials provide fast prediction of potential energy surfaces, closely matching the accuracy of the electronic structure methods used to produce the training data. However, NN predictions are only reliable within well-learned training domains, and show volatile behavior when extrapolating. Uncertainty quantification approaches can flag atomic configurations
    for which prediction confidence is low, but arriving at such uncertain regions requires expensive sampling of the NN phase space, often using atomistic simulations. Here, we exploit automatic differentiation to drive atomistic systems towards high-likelihood, high-uncertainty configurations without the need for molecular dynamics simulations. By performing adversarial attacks on an uncertainty metric, informative geometries that expand the training domain of NNs are sampled. When combined to an active learning loop, this approach bootstraps and improves NN potentials while decreasing the number of calls to the ground truth method. This efficiency is demonstrated on sampling of kinetic barriers and collective variables in molecules, and can be extended to any NN potential architecture and materials system.
\end{abstract}

\maketitle

\section{Introduction}

Recent advances in machine learning (ML) techniques have enabled the study of increasingly larger and more complex materials systems.\cite{Butler2018,Zunger2018,Schwalbe-Koda2020} In particular, ML-based atomistic simulations have demonstrated predictions of potential energy surfaces (PESes) with accuracy comparable to \textit{ab initio} simulations while being orders of magnitude faster.\cite{Behler2011a,Botu2017,Mueller2020} ML potentials employing kernels or Gaussian processes have been widely used for fitting PESes,\cite{Bartok2017,Chmiela2018,Vandermause2020b} and are particularly effective in low-data regimes. For systems with greater diversity in chemical composition and structures, such as molecular conformations or reactions, larger training datasets are typically needed. Neural networks (NNs) can fit interatomic potentials to extensive datasets with high accuracy and lower training and inference costs.\cite{Liu2018WhenGPs,Behler2007} Over the last years, several models have combined different representations and NN architectures to predict PESes with increasing accuracy.\cite{Behler2007,Schutt2018b,Zhang2018a,Wang2018Coarse,Klicpera2020a} They have been applied to predict molecular systems,\cite{Jose2012,Morawietz2016b} solids,\cite{Artrith2016} interfaces,\cite{Natarajan2016} chemical reactions,\cite{Gastegger2015,Ang2020} kinetic events,\cite{Khaliullin2011} phase transitions\cite{Cheng2020} and many more.\cite{Behler2011a,Mueller2020}

Despite their remarkable capacity to interpolate between data points, NNs are known to perform poorly outside of their training domain\cite{Barrett2018Measuring,Xu2020neural} and may fail catastrophically for rare events, such as those occurring in atomistic simulations with large sizes or time scales not explored in the training data. Increasing the amount and diversity of the training data is beneficial to improve performance,\cite{Wang2020a,Ang2020} but there are significant costs associated to generating new ground-truth data points. Continuously acquiring more data and re-training the NN along a simulation may negate some of the acceleration provided by ML models. In addition, exhaustive exploration or data augmentation of the input space is intractable. Therefore, assessing the trustworthiness of NN predictions and systematically improving them is fundamental for deploying ML-accelerated tools to real world applications, including the prediction of materials properties.

Quantifying model uncertainty then becomes key, since it allows distinguishing new inputs that are likely to be informative and worth labeling with \textit{ab initio} simulations, from those close to configurations already represented in the training data. In this context, epistemic uncertainty --- the model uncertainty arising from the lack of appropriate training data --- is much more relevant to ML potentials than the aleatoric uncertainty, which arises from noise in the training data. Whereas ML-based interatomic potentials are becoming increasingly popular, uncertainty quantification applied to atomistic simulations is at earlier stages.\cite{Peterson2017b,Venturi2020} ML potentials based on Gaussian processes are Bayesian in nature, and thus benefit from an intrinsic error quantification scheme, which has been applied to train ML potentials on-the-fly\cite{Jinnouchi2019,Vandermause2020b} or to accelerate nudged elastic band (NEB) calculations.\cite{GarridoTorres2019} NNs do not typically handle uncertainty natively and it is common to use approaches that provide distributions of predictions to quantify epistemic uncertainty. Strategies such as Bayesian NNs,\cite{pmlr-v37-blundell15} Monte Carlo dropout\cite{pmlr-v48-gal16} or NN committees\cite{Politis1994,Clemen1989,Zhao2005} allow estimating the model uncertainty by building a set of related models and comparing their predictions for a given input. In particular, NN committee force fields have been used to control simulations,\cite{Chen2020IterativeForces} to inform sampling strategies\cite{Schran2020} and to calibrate error bars for computed properties.\cite{Imbalzano2021UncertaintySampling}

Even when uncertainty estimates are available to distinguish informative from uninformative inputs, ML potentials rely on atomistic simulations to generate new trial configurations. For example, it is common to perform molecular dynamics (MD) simulations with NN-based models to expand their training set in an active learning (AL) loop.\cite{Shapeev2020a,Wang2020a,Ang2020} MD simulations explore phase space based on the thermodynamic probability of the PES. Thus, in the best case, ML-accelerated MD simulations produce outputs correlated to the training set if the ML potential accurately reproduces the ground truth PES. However, exploring this region only provides incremental improvement to the potentials, which will still struggle to observe rare events. In the worst case, MD trajectories can be unstable when executed with an NN potential and sample unrealistic events that are irrelevant to the true PES, especially in early stages of the AL cycle when the NN training set is not representative of the overall configuration space. Gathering data from \textit{ab initio} MD prevents the latter issue, although at a higher computational cost. Even NN simulations need to sample very large amounts of low uncertainty phase space before stumbling upon uncertain regions. Some works avoid performing dynamic simulations, but still require forward exploration of the PES to find new training points.\cite{Lin2020AutomaticallyStrategy} Hence, one of the major bottlenecks for scaling up NN potentials is minimizing their extrapolation errors until they achieve self-sufficiency to perform atomistic simulations within the full phase space they will be used in, including handling rare events. Inverting the problem of exploring the configuration space with NN potentials would allow for a more efficient sampling of transition states and dynamic control.\cite{Noe2019,Wang2020b}

In this work, we propose an inverse sampling strategy for NN-based atomistic simulations by performing gradient-based optimization of a differentiable, likelihood-weighted uncertainty metric. Building on the concept of adversarial attacks from the ML literature,\cite{Szegedy2014a,Goodfellow2015a} new molecular conformations are sampled by backpropagating atomic displacements to find local optima that maximize the uncertainty of an NN committee while balancing thermodynamic likelihood. These new configurations are then evaluated using atomistic simulations (e.g. density functional theory or force fields) and used to retrain the NNs in an AL loop. The technique is able to bootstrap training data for NN potentials starting from few configurations, improve their extrapolation power, and efficiently explore the configuration space. The approach is demonstrated in several atomistic systems, including finding unknown local minima in a toy two-dimensional PES, improving kinetic barrier predictions for nitrogen inversion, increasing the stability of MD simulations in molecular systems, and sampling of collective variables in alanine dipeptide. This work provides a new method to explore potential energy landscapes without the need for brute-force \textit{ab initio} MD simulations to propose trial configurations.

\section{Theory}

\begin{figure*}[htb]
\includegraphics[width=\linewidth]{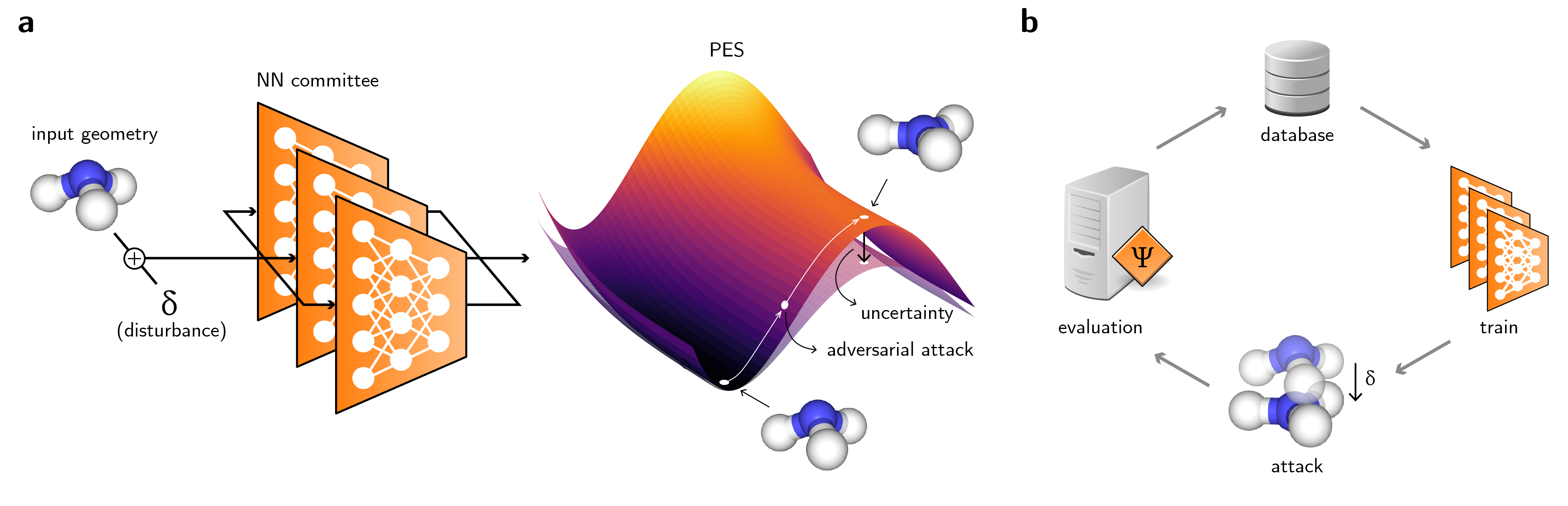}
\caption{\textbf{a}, Schematic diagram of the method. Nuclear coordinates of an input molecule are slightly displaced by $\delta$. Then, a potential energy surface (PES) and its associated uncertainty are calculated with an NN potential committee. By backpropagating an adversarial loss through the NN committee, the displacement $\delta$ can be updated using gradient ascent techniques until the adversarial loss is maximized, thus converging to states that compromise high uncertainty with low energy. \textbf{b}, Schematic diagram of the active learning loop used to train the NN potential committee. The evaluation can be performed with classical force fields or electronic structure methods.}
\label{fig:workflow}
\end{figure*}

\subsection{Neural network potentials}

An NN potential is a hypothesis function $\hyp$ that predicts a real value of energy $\hat{E} = \hyp (X)$ for a given atomistic configuration $X$ as input. $X$ is generally described by $n$ atoms with atomic numbers $\mathbf{Z} \in \Int^n$ and nuclear coordinates $\mathbf{R} \in \Real^{n \times 3}$. Energy-conserving atomic forces $F_{ij}$ on atom $i$ and Cartesian coordinate $j$ are obtained by differentiating the output energy with respect to the atomic coordinates $r_{ij}$,

\begin{equation}\label{eq:force_def}
    \hat{F}_{ij} = - \frac{\partial \hat{E}}{\partial r_{ij}}.
\end{equation}

\noindent The parameters $\theta$ are trained to minimize the expected loss $\Loss$ given the distribution of ground truth data $(X, E, \mathbf{F})$ according to the dataset $\mathcal{D}$,

\begin{equation}\label{eq:objective_1}
    \min_\theta \expected{(X, E, \mathbf{F}) \sim \mathcal{D}}{\Loss \left(X, E, \mathbf{F}; \theta\right)}.
\end{equation}

\noindent During training, the loss $\Loss$ is usually computed by taking the average mean squared error of the predicted and target properties within a batch of size $N$,

\begin{equation}\label{eq:loss}
    \Loss = \Mean{N}{i=1}{N} \left[ \alpha_E \norm{E_i - \hat{E}_i}^2 + \alpha_F  \norm{\mathbf{F}_i - \hat{\mathbf{F}}_i}^2 \right],
\end{equation}

\noindent where $\alpha_E$ and $\alpha_F$ are coefficients indicating the trade-off between energy and force-matching during training.\cite{Schutt2018b} The training proceeds using stochastic gradient descent-based techniques.

\subsection{Uncertainty quantification}

To create a differentiable metric of uncertainty, we turned to NN committees. These are typically implemented by training different $\hyp$ and obtaining a distribution of predictions for each input $X$. For example, given $M$ models implementing $\hat{E}^{(m)} = \hyp^{(m)}(X)$, the mean and the variance of the energy of an NN potential ensemble can be computed as

\begin{align}
    \bar{E}(X) &= \Mean{M}{m=1}{M} \hat{E}^{(m)} (X), \\
    \Var{E}(X) &= \Mean{M - 1}{m=1}{M} \norm{\hat{E}^{(m)} (X) - \bar{E} (X)}^2,
\end{align}
    
\noindent and similarly for forces,

\begin{align}
    \bar{\mathbf{F}}(X) &= \Mean{M}{m=1}{M} \hat{\mathbf{F}}^{(m)} (X), \\
    \Var{F}(X) &= \Mean{M - 1}{m=1}{M} \left[\Mean{3n}{i, j}{} \norm{\hat{F}^{(m)}_{ij} (X) - \bar{F}_{ij} (X)}^2 \right].
\end{align}

\noindent Whereas the training objective \eqref{eq:objective_1} rewards approaching mean energies or forces to their ground truth values, this is not guaranteed for regions outside of the training set.

Since variances in properties may become higher when the NN models are in the extrapolation regime, identifying whether an NN committee is outside its fitting domain requires evaluating the probability that the output of the NN is reliable for an input $X$. One option is to model this problem for the epistemic error as a simple classifier,

\begin{equation}\label{eq:extrapolation_prob}
    P(X \in \mathcal{D} \, |\, \Var{}) = 
    \begin{cases}
        1, \, \Var{} < t, \\
        0, \, \Var{} \geq t, \\
    \end{cases}
\end{equation}

\noindent with $t$ a threshold chosen by evaluating the model on the training set. Although Eq. \eqref{eq:extrapolation_prob} can be modified to accept the data $X$ with a certain likelihood, the deterministic classifier demonstrates reasonable accuracy (see Supporting Information for details).

\subsection{Adversarial training}

When developing adversarially robust models, the objective \eqref{eq:objective_1} is often changed to include a perturbation $\delta$,\cite{tsipras2018robustness}

\begin{equation}\label{eq:objective_robust}
    \min_\theta \expected{(X, E, \mathbf{F}) \sim \mathcal{D}}{\max_{\delta \in \Delta} \Loss \left(X_\delta, E_\delta, \mathbf{F}_\delta; \theta\right)},
\end{equation}

\noindent with $\Delta$ the set of allowed perturbations and $X_\delta, E_\delta, \mathbf{F}_\delta$ the perturbed geometries and their corresponding energies and forces, respectively. In the context of NN classifiers, $\Delta$ is often chosen as the set of $\ell_p$-bounded perturbations for a given $\varepsilon$, $\Delta = \{\delta \in \Real \; | \; \norm{\delta}_p \leq \varepsilon \}$. Adversarial examples are then constructed by keeping the target class constant under the application of the adversarial attack.\cite{Szegedy2014a,Goodfellow2015a} On the other hand, adversarial examples are not well defined for NN regressors. Since even slight variations of the input lead to different ground truth results $E_\delta, \mathbf{F}_\delta$, creating adversarially robust NN regressors is not straightforward.

We propose that creating adversarially-robust NN potentials can be achieved by combining adversarial attacks, uncertainty quantification, and active learning. Although similar strategies have been used in classifiers, graph-structured data,\cite{Zugner2018AdversarialData,Zhu2019RobustAttacks} and physical models,\cite{Cubuk2020AdversarialModels} no work has yet connected these strategies to sample potential energy landscapes. In this framework, an adversarial attack maximizes the uncertainty in the property under prediction (Fig. \ref{fig:workflow}a). Then, ground-truth properties are generated for the adversarial example. This could correspond to obtaining energies and forces for a given conformation with density functional theory (DFT) or force field approaches. After acquiring new data points, the NN committee is retrained. New rounds of sampling can be performed until the test error is sufficiently low or the phase space is explored to a desirable degree. Fig. \ref{fig:workflow}b illustrates this loop.

Within this pipeline, new geometries are sampled by performing an adversarial attack that maximizes an adversarial loss such as

\begin{equation}\label{eq:objective_attack}
    \max_{\delta \in \Delta} \LossAdv(X, \delta; \theta) = \max_{\delta \in \Delta} \Var{F}(X_\delta).
\end{equation}

\noindent In force-matching NN potentials, the uncertainty of the force may be a better descriptor of epistemic error than uncertainty in energy\cite{klicpera2020fast} (see Supporting Information).

In the context of atomistic simulations, the perturbation $\delta$ is applied only to the nuclear coordinates, $X_\delta = (\mathbf{Z}, \mathbf{R} + \bm{\delta})$,  $\bm{\delta} \in \Real^{n \times 3}$. For systems better described by collective variables (CVs) $\mathbf{s} = \mathbf{s} (\mathbf{R})$, an adversarial attack can be applied directly to these CVs, $X_\delta = (\mathbf{Z}, \mathbf{s}^{-1}(\mathbf{s} + \bm{\delta}))$, as long as there are some differentiable functions $\mathbf{s}^{-1}$ backmapping $\mathbf{s}$ to the nuclear coordinates $\mathbf{R}$.

The set $\Delta$ can be defined by appropriately choosing $\varepsilon$, the maximum $p$-norm of $\delta$. However, in atomistic simulations, it is often interesting to express these limits in terms of the energy of the states to be sampled, and the sampling temperature. To that end, a partition function $Q$ of the system at a given temperature $T$ can be constructed from the ground truth data $\mathcal{D}$,

\begin{equation}\label{eq:partition_fn}
    Q = \displaystyle{\sum_{(X, E, \mathbf{F}) \in \mathcal{D}}} \exp\left(-\frac{E}{kT}\right),
\end{equation}

\noindent with $k$ being the Boltzmann constant. Disregarding the entropic contributions, the probability $p$ that a state $X_\delta$ with predicted energy $\bar{E}(X_\delta)$ will be sampled is

\begin{equation}\label{eq:probability}
    p(X_\delta) = \frac{1}{Q} \exp\left(-\frac{\bar{E}(X_\delta)}{kT}\right).
\end{equation}

\noindent Finally, instead of limiting the norm of $\delta$, the adversarial objective can be modified to limit the energy of sampled states by combining Eqs. \eqref{eq:objective_attack} and \eqref{eq:probability},

\begin{equation}\label{eq:objective_attack_prob}
    \max_{\delta} \LossAdv(X, \delta; \theta) = \max_{\delta} p(X_\delta) \Var{F}(X_\delta).
\end{equation}

\noindent Using automatic differentiation strategies, the value of $\delta$ can be obtained by iteratively using gradient ascent techniques,

\begin{equation}\label{eq:adv_gradient}
    \delta^{(i+1)} = \delta^{(i)} + \alpha_\delta \frac{\partial \LossAdv}{\partial \delta},
\end{equation}

\noindent with $i$ the number of the iteration and $\alpha_\delta$ the learning rate for the adversarial attack.

In practice, adversarial examples require input geometries as seeds, and an appropriate initialization of $\bm{\delta}$. One possibility is to sample the initial $\bm{\delta}$ from a normal distribution $\mathcal{N}\left(0, \Var{\delta}\, \mathbf{I }\right)$ with a small value of $\Var{\delta}$. The degenerate case $\Var{\delta} = 0$ leads to deterministic adversarial attacks with the optimization procedure.

Since one can create several adversarial examples per initial seed, the computational bottleneck becomes evaluating them to create more ground truth data. Hence, reducing the number of adversarial examples is of practical consideration. Generated examples can be reduced by using only a subset of the initial dataset $\mathcal{D}$ as seeds. Even then, the optimization of $\delta$ may lead to structures which are very similar, corresponding to the same points in the configuration space. To avoid evaluating the same geometry multiple times, structures can be deduplicated according to the root mean square deviation (RMSD) between the conformers. One efficient algorithm is to perform hierarchical clustering on the data points given the RMSD matrix, and aggregating points which are within a given threshold of each other. Finally, to avoid local minima around the training set, one can classify whether the given structure is well-known by the model using Eq. \eqref{eq:extrapolation_prob}. Then, new points are evaluated only if they correspond to high uncertainty structures and not just to local optima in uncertainty, avoiding sampling regions of the PES which are already well represented in the training set.

The complete adversarial training procedure is described in Algorithm \ref{alg:1}.

\begin{algorithm}[htb]\label{alg:1}
\DontPrintSemicolon

    \KwInput{dataset $\mathcal{D}_1$, hyperparameters}
    \KwData{geometries, energies and forces $(X, E, \mathbf{F})$}

    \For{generation $g \in (1, ..., G)$}
    {
        \tcc{Training}
        \For{model $m \in (1, ..., M)$}    
        { 
        	sample a train set from $\mathcal{D}_g$ \;
        	train $\hyp^{(m)}$ to minimize $\Loss$  \tcp*{Eqs. \eqref{eq:objective_1},\eqref{eq:loss}}
        }
        
        \tcc{Adversarial Attack}
        construct $Q$ from $\mathcal{D}_g$ \tcp*{Eq. \eqref{eq:partition_fn}} \;
        sample attack seeds $\{X_i\}$ from $\mathcal{D}_g$ \;
        \For{seed $X_i \in \{X_i\}$}
        { 
        	initialize $\bm{\delta} \sim \mathcal{N}\left(0, \Var{\delta}\, \mathbf{I }\right)$ \;
        	train $\bm{\delta}$ to maximize $\LossAdv$
        	\tcp*{Eqs. \eqref{eq:objective_attack_prob},\eqref{eq:adv_gradient}} \;
        	$X_{\delta,i} := (\mathbf{Z}_i, \mathbf{R}_i + \bm{\delta})$
        }
        
        \tcc{Evaluating}
        
        initialize $\mathcal{D}_{g + 1}$ := $\mathcal{D}_{g}$ \;
        \For{adversarial example $X_{\delta,i}\in \{X_{\delta,i}\}$}
        { 
        	obtain ground-truth $\left(E_{\delta,i}, \mathbf{F}_{\delta,i}\right)$ for $X_{\delta,i}$ \;
        	add $\left(X_{\delta,i}, E_{\delta,i}, \mathbf{F}_{\delta,i}\right)$ to $\mathcal{D}_{g + 1}$
        }
    }

\caption{Adversarial training of a neural network potential}
\end{algorithm}

\section{Results and Discussion}

\subsection{Double well potential}\label{sec:2dwell}

\begin{figure*}[htb]
\includegraphics[width=\linewidth]{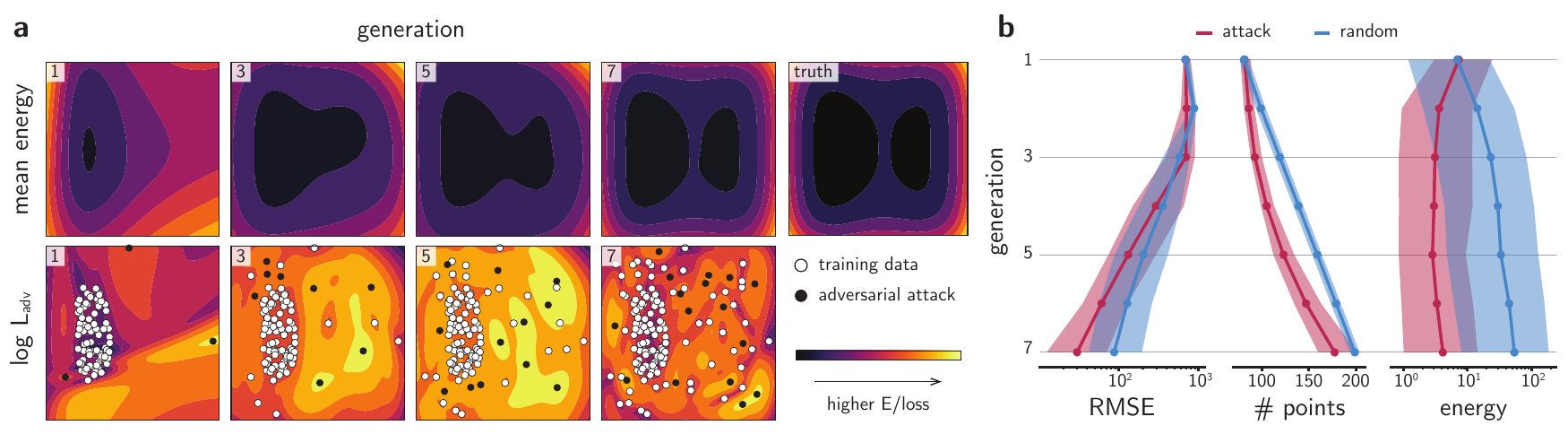}
\caption{\textbf{a}, Evolution of the PES of a 2D double well predicted by an NN committee. Adversarial examples (black dots) are distortions from original training data (white dots) or past adversarial examples (gray dots) that maximize the adversarial loss $\LossAdv$. The plotting intervals are $[-1.5, 1.5] \times [-1.5, 1.5]$ for all graphs. The generation of the NN committee is shown in the top left corner of each graph. \textbf{b}, Evolution of the root mean square error (RMSE), number of training points and energy of the sampled points for adversarial attack strategy (red) and randomly distorting the training data (blue). The solid line is the median from more than 100 experiments, and the shaded area is the interquartile region.}
\label{fig:well}
\end{figure*}

As a proof-of-concept, the adversarial sampling strategy is demonstrated in the two-dimensional (2D) double well potential (see Supporting Information and Figs. S1-S4 for an analysis of the 1D example). To investigate the exploration of the phase space, the initial data is placed randomly in one of the basins of the potential. Then, a committee of feedforward NNs is trained to reproduce the potential using the training data (see Methods). At first, the NN potential is unaware of the second basin, and predicts a single well potential in its first generation. As such, an MD simulation using this NN potential would be unable to reproduce the free energy surface of the true potential. Nevertheless, the region corresponding to the second basin is of high uncertainty when compared to the region where the training set is located. The adversarial loss encourages exploring the configuration space away from the original data, and adversarial samples that maximize $\LossAdv$ are evaluated with the ground truth potential, then added to the training set of the next generation of NN potentials. Fig. \ref{fig:well}a shows the results of the training-attacking loop for the NN potential after several generations. As the AL loop proceeds, the phase space is explored just enough to accurately reproduce the 2D double well, including the energy barrier and the shape of the basins.

To verify the effectiveness of the adversarial sampling strategy, the evolution of the models is compared with random sampling. While the former is obtained by solving Eq. \eqref{eq:objective_attack_prob}, the latter is obtained by randomly selecting 20 different training points from the training set and sampling $\delta$ from a uniform distribution, $\delta \sim \mathcal{U}\left(-\sigma_\delta, \sigma_\delta \right)$. To perform a meaningful statistical analysis on the methods, more than 100 independent active learning loops with different initializations are trained for the same 2D well potential (Fig. \ref{fig:well}b). Overall, the root mean square error (RMSE) between the ground truth potential and the predicted values decreases as the space is better sampled for both methods. However, although the random sampling strategy collects more data points, the median RMSE of the final generation is between two to three times higher than the adversarial attack strategy. Moreover, the median sampled energy is one order of magnitude higher for randomly-sampled points. As several randomly-sampled points travel to places outside of the bounds of the double well shown in Fig. \ref{fig:well}a, the energy quickly increases, leading to high-energy configurations. This is often the case in real systems, in which randomly distorting molecules or solids rapidly lead to high-energy structures that will not be visited during production simulations. As such, this toy example suggests that the adversarial sampling method generates thermodynamically likely structures, requires less ground-truth evaluations and leads to better-trained NN potentials compared to randomly sampling the space.

\subsection{Nitrogen inversion on ammonia}

\begin{figure*}[htb]
\includegraphics[width=\linewidth]{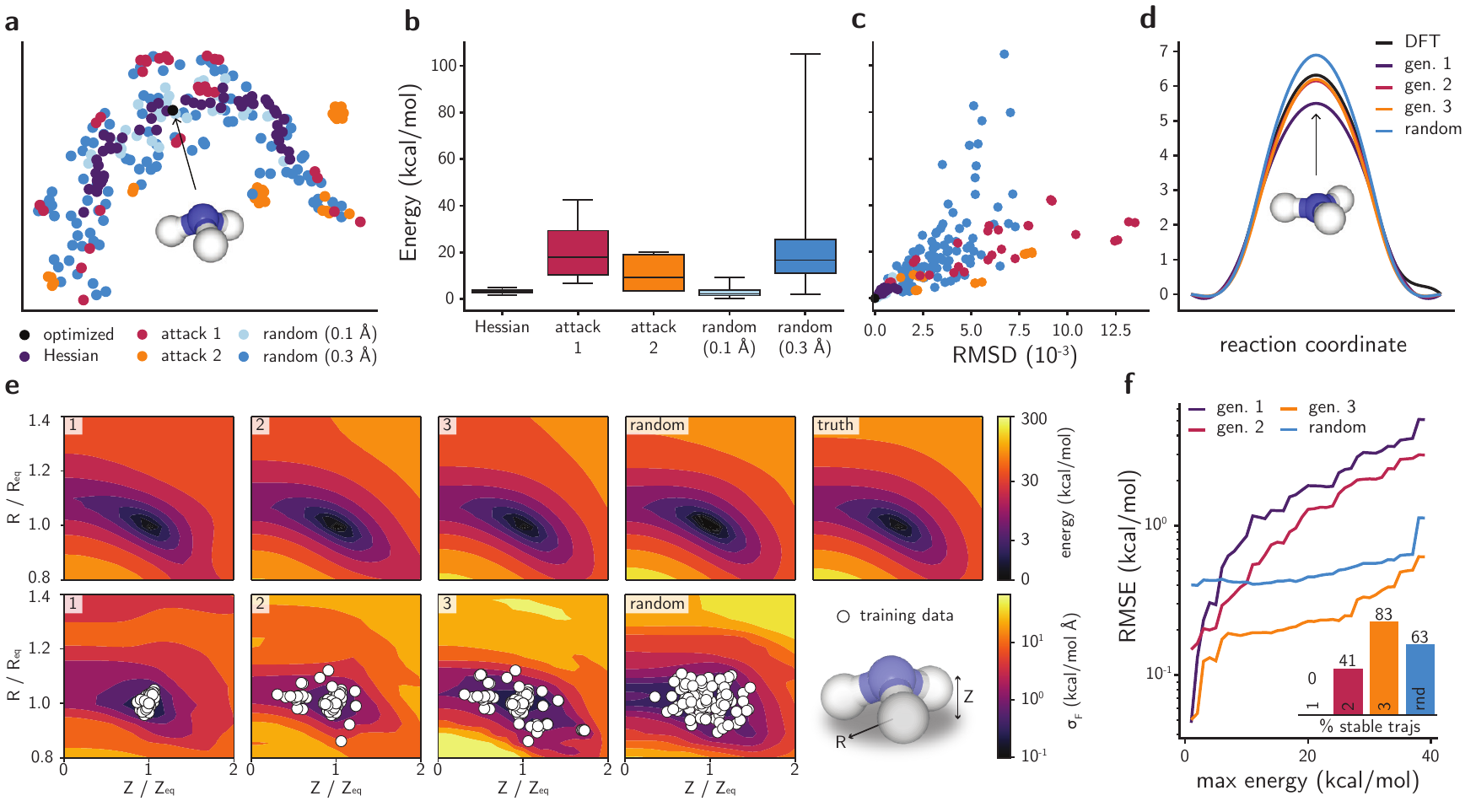}
\caption{\textbf{a}, UMAP plot for the SOAP-based similarity between ammonia geometries. Both axes are on the same scale. \textbf{b}, Distribution of DFT energies for conformations sampled with different methods. The horizontal line is the median, the box is the interquartile region and the whiskers span the range of the distribution. \textbf{c}, Relationship between DFT energy and root mean square deviation (RMSD) of a geometry with respect to the ground state structure of ammonia. The color scheme follows the legend of \textbf{a}. \textbf{d}, Energy barrier for the nitrogen inversion calculated with NEB using DFT or using the NN committee. \textbf{e}, Evolution of the PES projected onto the CVs ($Z$, $R$) for ammonia. The generation of the NN committee is shown in the top left corner of each plot. The scale bar of energies is plotted with the function $\log_{10} (1 + E)$, and all energy contour plots have the same levels. Random geometries were generated with $\sigma_\delta = 0.3$ \AA~ (see Methods). \textbf{f}, RMSE between the NN and DFT PES for each NN potential when a maximum energy is imposed for the DFT PES. \textbf{f, inset}, fraction of stable MD trajectories generated using each NN committee as force field.}
\label{fig:ammonia}
\end{figure*}

As a second example, we bootstrap an NN potential to study the nitrogen inversion in ammonia. This choice of molecule is motivated by more complex reactive systems, in which quantifying energy barriers to train a robust NN potential requires thousands of reactive trajectories from \textit{ab initio} simulations.\cite{Ang2020} To circumvent that need, we start training an NN committee using the SchNet model\cite{Schutt2018b} from Hessian-displaced geometries data. Then, new geometries are sampled by performing an adversarial attack on the ground state conformation, and later evaluated using DFT. After training a new committee with newly-sampled data points, the landscape of conformations is analyzed and compared with random displacements. Fig. \ref{fig:ammonia}a shows a UMAP visualization\cite{McInnes2018} of the conformers, as compared by their similarity using the Smooth Overlap of Atomic Positions (SOAP) representation.\cite{Bartok2013} A qualitative analysis of the UMAP plot shows that adversarial attacks rarely resemble the training set in terms of geometric similarity. Attacks from the second generation are also mostly distant from attacks in the first generation. On the other hand, small values of distortions $\sigma_\delta$ for a uniform distribution (see Sec. \ref{sec:2dwell})  create geometries that are very similar to Hessian-displaced ones. While higher values of $\sigma_\delta$ (e.g. $\sigma_\delta = 0.3$ \AA) explore a larger conformational space, several points with very high energy are sampled (Fig. \ref{fig:ammonia}b), as in the double well example. As the number of atoms increases, this trade-off between thermodynamic likelihood and diversity of the randomly sampled configurations worsens in a curse-of-dimensionality effect. In contrast, energies of adversarially created conformations have a more reasonable upper bound if the uncertainty in forces is used. When the uncertainty in energy is employed in Eq. \eqref{eq:objective_attack} instead of $\Var{F}$, adversarial examples may not efficiently explore the configuration space (Fig. S5), supporting the use of $\Var{F}$ for performing inverse sampling. Although calculating gradients with respect to $\Var{F}$ requires more memory to store the computational graph (Fig. S6), this metric is more informative of epistemic uncertainty in NN potentials than its energy counterpart (Figs. S7, S8) and better reflects the preference of force-matching over energy-matching at train time.  Fig. \ref{fig:ammonia}c compares the degree of distortion of the geometries with respect to their energies. It further shows that the adversarial strategy navigates the conformational space to find highly distorted, lower energy states. Both the first and second generation of attacked geometries display higher RMSD than Hessian-displaced structures with respect to the ground state geometry while staying within reasonable energy bounds. However, as the low-energy region of the PES is better explored by the NN potential as the AL loop progresses, adversarially sampled geometries from later generations become increasingly higher in energy (Fig. S9).

Once new configurations are used in training, predictions for the energy barrier in the nitrogen inversion improve substantially (Fig. \ref{fig:ammonia}d). While the first generation of the NN potential underestimates the energy barrier by about 1 kcal/mol with respect to the DFT value, the prediction from the second generation is already within the error bar, with less than 0.25 kcal/mol of error for the inversion barrier (see Fig. S10). In contrast, predictions from an NN committee trained on randomly-sampled geometries overestimate this energy. They also exhibit higher uncertainties, even for geometries close to equilibrium (Fig. S10). This suggests that the adversarial attack was able to sample geometries similar to the transition state of the nitrogen inversion reaction and accurately interpolate the energy barrier without the need to manually add this reaction path into the training set.

The evolution of the phase space of each NN committee is further compared in the projected PES of Fig. \ref{fig:ammonia}e (see Methods). Two CVs are defined to represent the phase space of this molecule: the radius of the circumference defined by the three hydrogen atoms ($R$) and the distance between the nitrogen atom and the plane defined by the three hydrogens ($Z$) (Fig. S11). Fig. \ref{fig:ammonia}e shows the energies and force uncertainties calculated for the most symmetrical structures containing these CVs (see Fig. S11a), with $R$, $Z$ normalized by the values found in the ground state geometry. Analogously to Fig. \ref{fig:well}a, adversarial attacks expand the configuration space used as train set for NN committees and bring the phase space closer to the ground truth, thus lowering the uncertainty of forces in the phase space (see also Figs. S7, S8). Nevertheless, randomly-sampled geometries also allow bootstrapping an NN committee depending on the system and values of $\sigma_\delta$. Importantly, NN committees successively trained on adversarial attacks have smaller errors in the low-energy region of the PES of ammonia. As expected, the high-energy configurations sampled by randomly-generated geometries slightly improve the higher energy region of the PES that will not be visited in production simulations. Fig. \ref{fig:ammonia}f shows the RMSE of each model compared to DFT across all the projected phase space of Fig. \ref{fig:ammonia}e. When only energies smaller than 5 kcal/mol are compared, all three generations display much smaller RMSE than random, probably due to the presence of Hessian-displaced geometries in their training set. The third generation of NN committees has a smaller RMSE when compared to random up to 40 kcal/mol, further supporting that the adversarial sampling strategy is useful to balance exploration of diverse conformations with higher likelihood. Finally, the adversarial training yields models capable of performing stable MD simulations. Whereas the first generation cannot produce stable MD trajectories, i.e. always leading to unphysical configurations such as atomic dissociation or collision, 83\% of the trajectories produced by the third generation of adversarially based NN committees are stable, even though the NN-based MD geometries include data points originally not in the training set (Fig. S12). In contrast, only 63\% of the trajectories are stable when the NN committee trained on random geometries is used. This indicates that the adversarial sampling strategy enhances the robustness of NN-based MD simulations by seeking points which are known to cause instabilities due to extrapolation errors, and unlikely to exist in training sets created by unbiased MD simulations (Fig. S12).

\subsection{Collective variable sampling in alanine dipeptide}

\begin{figure*}[htb]
\includegraphics[width=\linewidth]{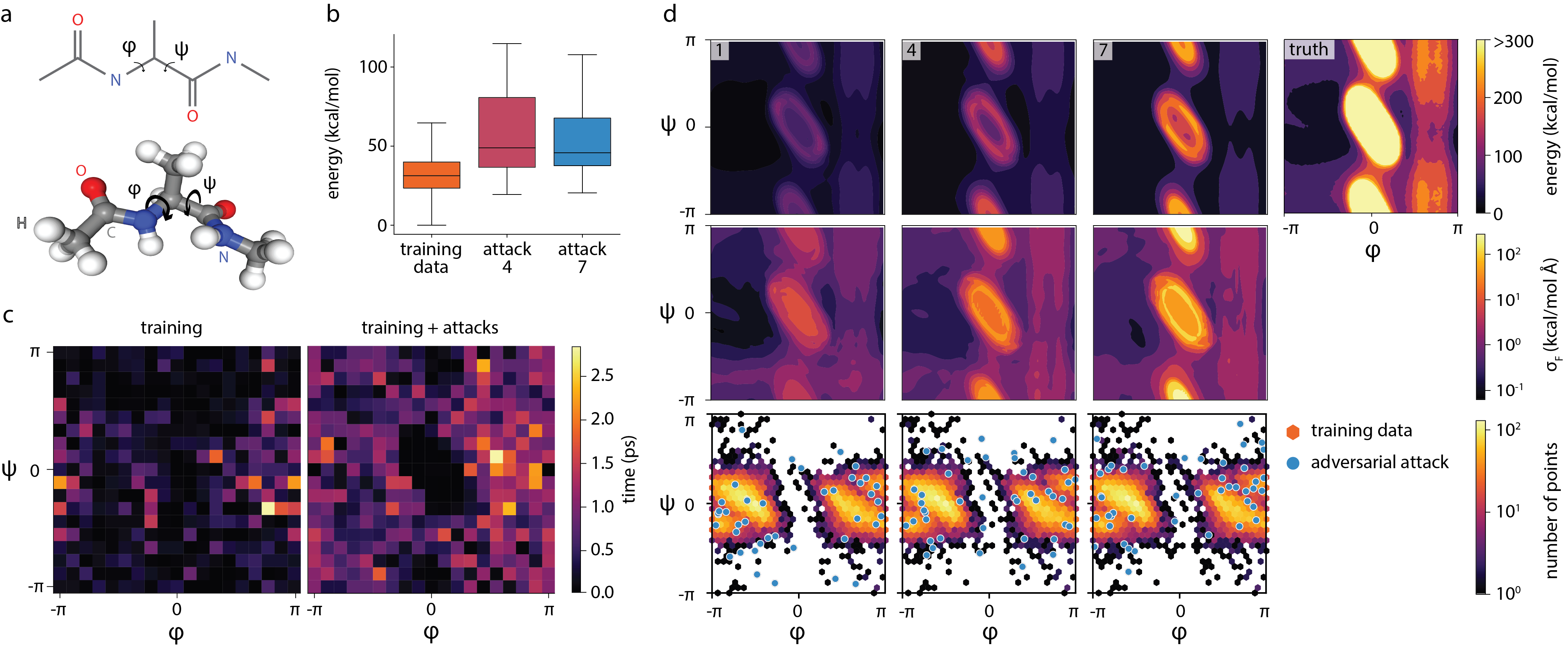}
\caption{\textbf{a}, Schematic diagram of alanine dipeptide and the CVs ($\varphi$, $\psi$) created from the highlighted dihedral angles. Hydrogen, carbon, oxygen and nitrogen atoms are depicted with white, gray, red and blue spheres, respectively. \textbf{b}, Distribution of force field energies for conformations generated from the collective variables. The horizontal line is the median, the box is the interquartile region and the whiskers span the range of the distribution. \textbf{c}, Duration of stable alanine dipeptide trajectories simulated with NN committees trained without and with adversarial examples. Grid points represent dihedral angles ($\varphi, \psi$) of the starting configurations of trajectories obtained via rigid rotation of the lowest energy geometry. \textbf{d}, Evolution of the PES of an NN committee trained on alanine dipeptide. Adversarial examples (red points) are distortions along the CVs ($\varphi$, $\psi$) from randomly chosen original training points (hexagonal bins) through Eq. \eqref{eq:objective_attack_prob}. Angles are given in radians.}
\label{fig:alanine}
\end{figure*}

As a final example, we illustrate the use of adversarial attacks for sampling predefined CVs. Since translation-based adversarial attacks $X_\delta = (\mathbf{Z}, \mathbf{R} + \bm{\delta})$ may not be able to capture collective dynamics of interest such as bond rotations (see full discussion on the Supporting Information), we seek to find high-uncertainty conformations in predefined CVs $\mathbf{s} = \mathbf{s} (\mathbf{R})$. To do that, there should exist a differentiable function $\mathbf{s}^{-1}$ mapping a point in the CV space to the atomic coordinate space $\Real^{n \times 3}$. Typically, CVs aggregate information from many degrees of freedom and $\mathbf{s} (\mathbf{R})$ may not be bijective. Nevertheless, in the case of adversarial attacks, it suffices to have an operation $\mathbf{s}^{-1}$ that acts on a geometry $X = \left(\mathbf{Z}, \mathbf{R}\right)$ to produce the adversarial attack $X_\delta = \left(\mathbf{Z}, \mathbf{s}^{-1} \left(X, \bm{\delta} \right)\right)$. Using this strategy, a seed geometry can be distorted in the direction of its predefined CVs even if the CVs are not invertible.

This application is illustrated with the alanine dipeptide (N-acetyl-L-alanine-N'-methylamide) molecule, using its two dihedral angles ($\varphi$, $\psi$) as CVs (Fig. \ref{fig:alanine}a). Despite their apparent chemical simplicity, flexible molecules pose tremendous challenges to NN potentials,\cite{Vassilev-Galindo2021ChallengesMolecules}, which are typically benchmarked on molecules with barely any rotatable bonds (e.g. MD17). In this particular case, the function $\mathbf{s}^{-1}$ takes a reference geometry $X$ as an input and performs the dihedral rotations of interest through purely geometrical operations. Since bond rotations can be written with matrix operations, they can be implemented in the training pipeline without breaking the computational graph that enables the adversarial strategy. To study the effects of the adversarial learning method, a series of NN committees were trained using the same architecture employed in the previous section. The models were trained on geometries created from MD simulations using the Optimized Potentials for Liquid Simulations (OPLS) force field\cite{Robertson2015}  with the OpenMM package\cite{Friedrichs2009,Eastman2017} (see Methods). Then, adversarial attacks were performed by randomly taking training points as seed geometries. Since bond rotations are periodic, the adversarial distortion $\delta$ does not break the geometries apart, a concern that existed in the previous section or in many other ML-accelerated simulations. Nevertheless, some angles ($\varphi$, $\psi$) may lead to high energy configurations depending on the conformation $X$ of the molecule prior to the attack. Fig. \ref{fig:alanine}b shows the distribution of sampled energies for different rounds of adversarial attacks. We discarded points with extremely high energy from the training set, since they interfere with the training of the NN potential for being overly far from equilibrium. Nevertheless, the distribution of energies show that most of the sampled points lie in energy ranges that are not accessible by unbiased short MD simulations but are reasonably expected to be accessed in production simulations. This further supports the hypothesis that adversarial attacks are effective in sampling regions of the phase space with good compromise between energy and uncertainty. To confirm that the adversarial sampling strategy improves the robustness of the NN potential, the stability of MD trajectories is computed for various initial configurations. Fig. \ref{fig:alanine}c compares the duration of stable trajectories obtained with the first and seventh generation of NN committees. As expected, the first generation produces very unstable trajectories, as even nanoseconds of MD simulations do not provide enough data to stabilize the NN potential. On the other hand, adding a relatively small number of adversarial examples enhances the robustness of the NN committees, as reflected in more stable MD trajectories (see also Fig. S13). Furthermore, since high energy adversarial points are discarded from the training, the NN committee is unable to produce stable trajectories for starting configurations with CVs near $(\varphi, \psi) = (0, 0)$.

The evolution of NN committees for predicting the PES of alanine dipeptide is shown in Fig. \ref{fig:alanine}d. At first, only a small region of the phase space is known from the data obtained in MD simulations. This is reflected on the high contrast between the uncertainty close and far from the training set. In the first few adversarial attacks, the space is better sampled according to the uncertainty metric, decreasing the uncertainty for $|\psi| > 0$ and increasing the uncertainty in high-energy regions. This suggests that the quality of the epistemic error classification improves as the conformation space is better explored. To better compare the ground truth results with the NN predictions in the low-energy region, we clipped the energies of the former to 300 kcal/mol in Fig. \ref{fig:alanine}d. As the active learning loop progresses, the NN committee is able to better reproduce the energy landscape of alanine dipeptide, as exemplified by the improvement of the CV landscape for $\varphi > \pi / 2$ or the high-energy ellipsoid centered at $(\varphi, \psi) = (0, 0)$, which will not be visited in unbiased simulations. Interestingly, the uncertainty remains high in the central region, since the sampled energies of the system are much higher than the rest of the phase space. Since some of them are discarded for being extremely unlikely (e.g. configurations with energies greater than $200$ kcal/mol), the predictive power of the NN committee is not guaranteed in  this part of the phase space. This is characterized by the ring-like energy barrier featured in Fig. \ref{fig:alanine}d for later generations. It is unclear whether NN potentials are able to simultaneously predict ground state conformations and such high energy states with similar absolute accuracy.\cite{Vassilev-Galindo2021ChallengesMolecules} In fact, learning high energy regions of the PES may not be needed, since the learned barriers are insurmountable in production simulations. Finally, the uncertainty in forces resembles traditional biasing potentials in enhanced sampling techniques applied to obtain the free energy landscape in alanine dipeptide.\cite{Laio2002,Zhang2019} Although this intuition is not thoroughly quantified in this work, we suggest that NN potentials with uncertainty quantification intrinsically provide a bias towards transition states through the uncertainty metric. Although the uncertainty can vary outside of the training set, as seen in Fig. \ref{fig:alanine}d, this idea  qualitatively agrees with the examples in this paper (see also Fig. S1a, S5). While we explore this bias through adversarial attacks for bootstrapping NN potentials in this work, we further suggest they could lead to automatic transition state and rare-event sampling strategies based on differentiable atomistic simulations with the uncertainty as a collective variable itself. This opportunity will be explored in future works.

\section{Conclusions}

In summary, we proposed a new sampling strategy for NN potentials by combining uncertainty quantification, automatic differentiation, adversarial attacks, and active learning. By maximizing the uncertainty of NN predictions through a differentiable metric, new geometries can be sampled efficiently and purposefully. This technique allows NN potentials to be bootstrapped with fewer calls to the ground truth method, maximizing the final accuracy and efficiently exploring the conformational space. Successful adversarial attacks were demonstrated in three examples. In a 2D double well potential, the attacks provided an exploration strategy and outperformed a random baseline. In the ammonia molecule, the approach accurately predicted distorted configurations or reaction paths, and produced better fits to the PES and more stable atomistic simulations, without the need of AIMD. For alanine dipeptide, a challenging molecule for NN potentials due to its flexibility, adversarial attacks were performed on collective variables to efficiently explore phase space and systemically improve the quality of the learned PES. This work presents a new data- and sample-efficient way to train NN potentials and explore the conformational space through deep learning-enabled simulations. By balancing thermodynamic likelihood and attacking model confidence it becomes possible to gather representative training data for uncertain, extrapolative configurations corresponding to rare events that would otherwise only be visited during expensive production simulations. The approach can be extended to any NN-based potential and representation, since it relies on the same backpropagation approach needed to train the potential by force-matching. 

\section{Methods}

\subsection{Double well potential}

The double well potential adopted in this work is written as the following polynomial:

\begin{equation}\label{eq:dw_equation}
    E(x, y) = 10 x^4 - 10 x^2 + 2x + 4 y^2.
\end{equation}

Initial training data was generated by randomly sampling up to 800 points with independent coordinates according to a uniform distribution $\mathcal{U}\left(-1.5, 1.5 \right)$, and selecting only those with energy lower than $-2$. This allows us to select only data points lying in the lowest energy basin of the double well, creating an energy barrier between the two energy minima.

Five feedforward NNs with four layers, softplus activation and 1,024 units per layer were trained using the same train/test splits of the dataset. The NNs had different initial weights. The dataset was split in the ratio 60:20:20 for training : validation : testing, with a batch size of 35. The training was performed for 600 epochs with the Adam optimizer\cite{Kingma2015} and a learning rate of 0.001. The reported RMSE is the mean squared difference between the average predicted energy $\bar{E}$ and the ground truth potential $E$ as evaluated on a $100 \times 100$ grid in the interval $[-1.5, 1.5] \times [-1.5, 1.5]$.

Adversarial attacks were performed with a normalized sampling temperature of 5 (Eq. \eqref{eq:dw_equation} units) for 600 epochs, learning rate of 0.003 and the Adam optimizer. Deduplication via hierarchical clustering was performed using a threshold of 0.02 for the distance and the 80th percentile of the train set variance.

Random distortions were performed in each generation by displacing the $(x, y)$ coordinates of training data points (or past random samples) by $\delta \sim \mathcal{U}\left(-1.0, 1.0 \right)$. After deduplication via hierarchical clustering and uncertainty percentile as performed for adversarial attacks, up to 20 points were randomly selected from the resulting data. Distortions smaller than 1.0 were often unable to efficiently explore the PES of the double well, landing in the same basin.

\subsection{Simulations of ammonia}\label{sec:methods_ammonia}

Initial molecular conformers were generated using RDKit \cite{Landrum2006} with the MMFF94 force field. \cite{Halgren1996} DFT structural optimizations and single-point calculations were performed using the BP86-D3/def2-SVP \cite{Becke1988, Perdew1986} level of theory as implemented in ORCA. \cite{Neese2018} NEB calculations\cite{Jonsson1998,Henkelman2000} were performed with 11 images using the FIRE algorithm\cite{Bitzek2006} as implemented in the Atomic Simulation Environment.\cite{HjorthLarsen2017} Hessian-displaced geometries were created by randomly displacing the atoms from their ground state conformation in the direction of normal mode vectors with temperatures between 250 and 750 K. In total, 78 training geometries were used as initial dataset.

For each generation, five NNs with the SchNet architecture\cite{Schutt2018b} were employed. Each model used four convolutions, 256 filters, atom basis of size 256, 32 learnable gaussians and cutoff of 5.0 \AA. The models were trained on different splits of the initial dataset (ratios $60 : 20 : 20$ for train : validation : test) for 500 epochs, using the Adam optimizer with an initial learning rate of $3 \times 10^{-4}$ and batch size of 30. A  scheduler reduced the learning rate by a factor of 0.5 if 30 epochs passed without improvement in the validation set. The training coefficients $\alpha_E$ and $\alpha_F$ (see Eq. \ref{eq:loss}) were set to 0.1 and 1, respectively.

Adversarial attacks were initialized by displacing the ground state geometry of ammonia by $\delta \sim \mathcal{N}(0, 0.01 \, \mathrm{\AA})$ for each coordinate. The resulting attack $\mathbf{\delta}$ was optimized for 60 iterations using the Adam optimizer with learning rate of 0.01. The normalized temperature $kT$ was set to 20 kcal/mol to ensure that adversarial attacks were not bound by a low sampling temperature, but by the uncertainty in force predictions. 30 adversarial attacks were sampled for each generation. No deduplication was performed.

Random distortions were generated by displacing each coordinate of the ground state geometry of ammonia by a value of $\delta \sim \mathcal{U}(-\sigma_\delta, \sigma_\delta)$. The values of $\sigma_\delta = 0.1$ \AA~ and $\sigma_\delta = 0.3$ \AA~ were adopted. 30 (100) random samples were created for $\sigma_\delta = 0.3$ \AA ($\sigma_\delta = 1.0$ \AA).

NN-based MD simulations were performed in the NVT ensemble with Nos\'e-Hoover dynamics, 0.5 fs timesteps, and temperatures of 500, 600, 700, 800, 900, and 1000 K. 100 5 ps-long trajectories were performed for each NN committee and temperature. The ground state geometry of ammonia was used as initial configuration for all MD calculations. Trajectories were considered as unphysical if the distance between hydrogen atoms was closer than 0.80 \AA or larger than 2.55 \AA, or if the predicted energy was lower than the ground state energy (0 kcal/mol for the reference adopted in this work).

SOAP vectors were created using the DScribe package.\cite{Himanen2020} The cutoff radius was set as 5 \AA, with spherical primitive Gaussian
type orbitals with standard deviation of 1 \AA, basis size of 5 functions, and $L_\mathrm{max} = 6$. The vectors were averaged over sites before summing the magnetic quantum numbers.

The projected PES shown in Fig. \ref{fig:ammonia}e is constructed by evaluating the NN potentials on symmetrical geometries generated for each tuple ($Z$, $R$). As such, train points and adversarial attacks are projected onto this space even though the conformers display distortions not captured by the CVs ($Z$, $R$) (see Fig. S11). The RMSE between the projected PES of the NN potential and DFT calculations is taken with respect to these symmetrical geometries.

\subsection{Simulations of alanine dipeptide}\label{sec:methods_alad}

Alanine dipeptide was simulated using the OPLS force field\cite{Robertson2015} within the OpenMM simulation package.\cite{Friedrichs2009,Eastman2017} The force field parameters were generated using LigParGen.\cite{Dodda2017} The molecule was placed in vacuum, with a box of size 30 \AA. MD simulations were performed at 1200 K using a Langevin integrator with a friction coefficient of 1 ps$^{-1}$ and step sizes of 2 fs. Calculations of Lennard-Jones and Coulomb interactions were performed in real space with no cutoff. The initial training data was obtained by conducting 160 2 ns-long MD simulations, from which snapshots every 2 ps were collected. 10,000 snapshots were extracted from these trajectories as the initial training data for the NN committee. 

For each generation, five NNs with the SchNet architecture\cite{Schutt2018b} were employed. The NNs follow the same architecture employed in the simulation of ammonia (see \ref{sec:methods_ammonia}). The models were trained for 200 epochs, using the Adam optimizer with an initial learning rate of $5 \times 10^{-4}$, batch size of 50, and learning rate scheduler. The training coefficients $\alpha_E$ and $\alpha_F$ (see Eq. \ref{eq:loss}) were both set to 1.0.

Adversarial attacks were initialized by displacing the CVs $(\varphi, \psi)$ by $\delta \sim \mathcal{N}(0, 0.01 \, \mathrm{rad})$ for each angle. The resulting attack $\mathbf{\delta}$ was optimized for 300 iterations using the Adam optimizer with learning rate of $5 \times 10^{-3}$. Normalized temperature of $kT$ was set to 20 kcal/mol. 50 adversarial attacks were sampled for each generation. No deduplication was performed. 

NN-based MD simulations were performed in the NVE ensemble using Velocity Verlet integrator at 300 K with a timestep of 0.5 fs. Trajectories starting from 324 different initial configurations were performed for each NN committee. Each starting geometry was obtained via rotation of the dihedral angles of the ground-state configuration while keeping the connected branches rigid. Trajectories were considered unstable if distance between the atoms are smaller than 0.75 \AA\ or larger than 2.0 \AA.

\section*{Acknowledgements}
D.S.-K. acknowledges the MIT Energy Fellowship for financial support. A.R.T. thanks Asahi Glass Company for financial support. R.G.-B. acknowledges support from ARPAe DIFFERENTIATE DE-AR0001220.

\section*{Author Contributions}

R.G.-B. supervised the research. D.S.-K. conceived the project. D.S.-K. and A.R.T. designed the experiments, performed the simulations, and wrote the computer code. All authors contributed to the data analysis and manuscript writing.

\section*{Competing Interests}

The authors declare no competing interests.

\section*{Data Availability}

The code used to reproduce the results from this paper is available at \url{https://github.com/learningmatter-mit/Atomistic-Adversarial-Attacks} under the MIT License.

\section*{References}



\section*{Supplementary material}

**Supporting Information for: Differentiable sampling of molecular geometries with uncertainty-based adversarial attacks**

Daniel Schwalbe-Koda,[*)] Aik Rui Tan,[*)] and Rafael Gómez-Bombarelli[†)]

*Department of Materials Science and Engineering, Massachusetts Institute of Technology, Cambridge, MA 02139*

(Dated: 28 March 2021)

---
[*)]D.S.-K. and A.R.T. contributed equally to this work
[†)]Electronic mail: rafagb@mit.edu

# I. SUPPLEMENTARY TEXT

## A. One-dimensional double well potential

To exemplify the training of the NN potential and the adversarial attacks, we analyze a simple, one-dimensional (1D) double well potential. At first, we employ a symmetrical potential described by

$$E(x) = 5x^4 - 10x^2. \tag{1}$$

Following the same methods employed for the 2D double well potential, we train the NN committee on ground truth data generated for this potential. Only data points $(r, E, F)$ with $E < -3.5$ are included in the train set. Fig. S1a shows the resulting prediction and train statistics for this system. The uncertainty in the energy increases for the barrier between the two wells, since no training data points are found in that region. While that is mostly the case for the variance of the forces, these also suffer from a drop around $r = 0$. This indicates that the networks agree that the force should go to zero near the point $r = 0$ to make the two wells coincide.

A statistical analysis of the uncertainties is shown in Fig. S1b. By taking data points with energy below zero which are not in the training set, we construct a test set where the NN ensemble is in the extrapolation regime. Then, we compare the variances of energies and forces for points in both the train and test sets. Overall, the distribution of variances in forces of the train set overlaps less with the test set distribution than their energy variance counterparts. This becomes clearer when the percentiles of the train (test) set are taken with respect to the variances of the test (train) distribution, as shown in Fig. S1c. In the case of force variances, most of the test set points are above the 70th percentile of the variances of the train set. Conversely, uncertainties of the train set are under the 20th percentile of the distribution of test set variances. This is not true for energy variances, where the test set has points with percentiles as low as 20th with respect to the train set.

Empirically, it can be seen that adopting the classification threshold $t$ (see Eq. (6) of the main text) as the 80th percentile of the forces variance from the train set is a reasonable choice. Fig. S1d shows the relationship between the mean absolute error (MAE) of energy and force predictions and the location of the 80th percentile of the variances. Whereas higher

errors are correlated to higher variances, a high variance does not necessarily imply a high error. This is often the case of undersampled regions where the interpolating power of the NN potential is able to perform good predictions. Moreover, errors are more pronounced for variances above the 80th percentile threshold for forces, illustrating its classification power for epistemic error.

Using the variance in forces $\sigma_F^2$ as the uncertainty metric, we construct the adversarial loss $\mathcal{L}_{\text{adv}}$ using Eq. (11) from the main paper, as shown in Fig. S2. In particular, Fig. S2d illustrates that upon a reasonable choice of temperature, transition states and points beyond the training set are favorably sampled by the adversarial attack.

To demonstrate the ability of adversarial attacks and active learning loops to efficiently explore the phase space, we repeat Fig. 2a of the main text for the 1D well (Fig. S3) by imposing an offset between the two wells,

$$E(x) = 5x^4 - 10x^2 + 1.5x. \qquad (2)$$

After the successive application of adversarial attacks, the well centered at $x = 1$ is discovered and sampled until the uncertainty in the region becomes below the 80th percentile of the force variance. The choice of temperature often prevents the adversarial loss from going towards infinity as $r \to \pm\infty$ unless the predicted energy becomes negative in these directions (generations 2 and 3 of Fig. S3). Sampling the points $r \to \pm\infty$ is avoided by using a fixed number of steps for optimizing $\delta$, which essentially limits how much the attack can travel.

If the energy uncertainty $\sigma_E^2$ is employed to create the adversarial loss $\mathcal{L}_{\text{adv}}$ instead of the force uncertainty, the exploration of the phase space is not performed as efficiently. Fig. S4 exemplifies an active learning loop for the 1D double well described by Eq. (2). Due to the force matching strategy when training the NN potential, the uncertainty outside of the training set is not necessarily higher than that within the training domains. As such, the adversarial attack on the uncertainty does not necessarily reward sampling new regions of the phase space, preventing the active learning loop from proceeding adequately every generation.

## B. Combining collective variable attacks and all-atom translations

In addition to performing adversarial attacks on predefined collective variables (CVs), we can also sample new geometries by applying small distortions ($\delta$) to the positions of all atoms, thus creating new geometries not seen in the training data. After training each generation of NN committee on the original molecular dynamics data (see Methods for NN training details), 700 training configurations from the training data are randomly chosen as seed geometries for the adversarial attacks. To ensure that physically meaningful configurations are obtained from the sampling, the normalized temperature $kT$ of the adversarial loss is set to 3 kcal/mol. The attacks were performed for 80 epochs using the Adam optimizer with a learning rate of $5 \times 10^{-3}$. This all-atom (AA) adversarial attack strategy was performed for 3 generations. Although the AA attack strategy allows the NN committee to sample various high-energy configurations, the phase space is not well explored, as CVs of attacked geometries are similar to the original seed CVs. In fact, the robustness of NN potential does not increase even when the size of the training data increases by around 20 % (Fig. S13).

To sufficiently explore the configuration space, we coupled 3 generations of AA attacks after 7 generations of CV attacks (CV + AA) (see Fig. S14). The CV adversarial attack follows the same procedure employed in the main paper (see Methods). After 7 generations of CV attacks, 3 generations of AA attacks with the procedure described above were performed. For each generation, 500 adversarial attacks were sampled. Half of the seed geometries was randomly selected from CV adversarial attacks while the other half was taken from the molecular dynamics data. While AA attacks alone fail to improve robustness of NN potentials, CV + AA attacks were able to yield much longer stable MD trajectories (Fig. S13). This suggests that CV attacks are necessary to capture collective dynamics such as bond rotations which are not easily explored via translation-based adversarial attacks. CV attacks allow the sampled configurations to access geometries outside the low-energy bounds. On the other hand, translation-based adversarial attacks supplement the diverse vibrational space offered by the increasing number of atoms in the system.

## II. SUPPLEMENTARY FIGURES

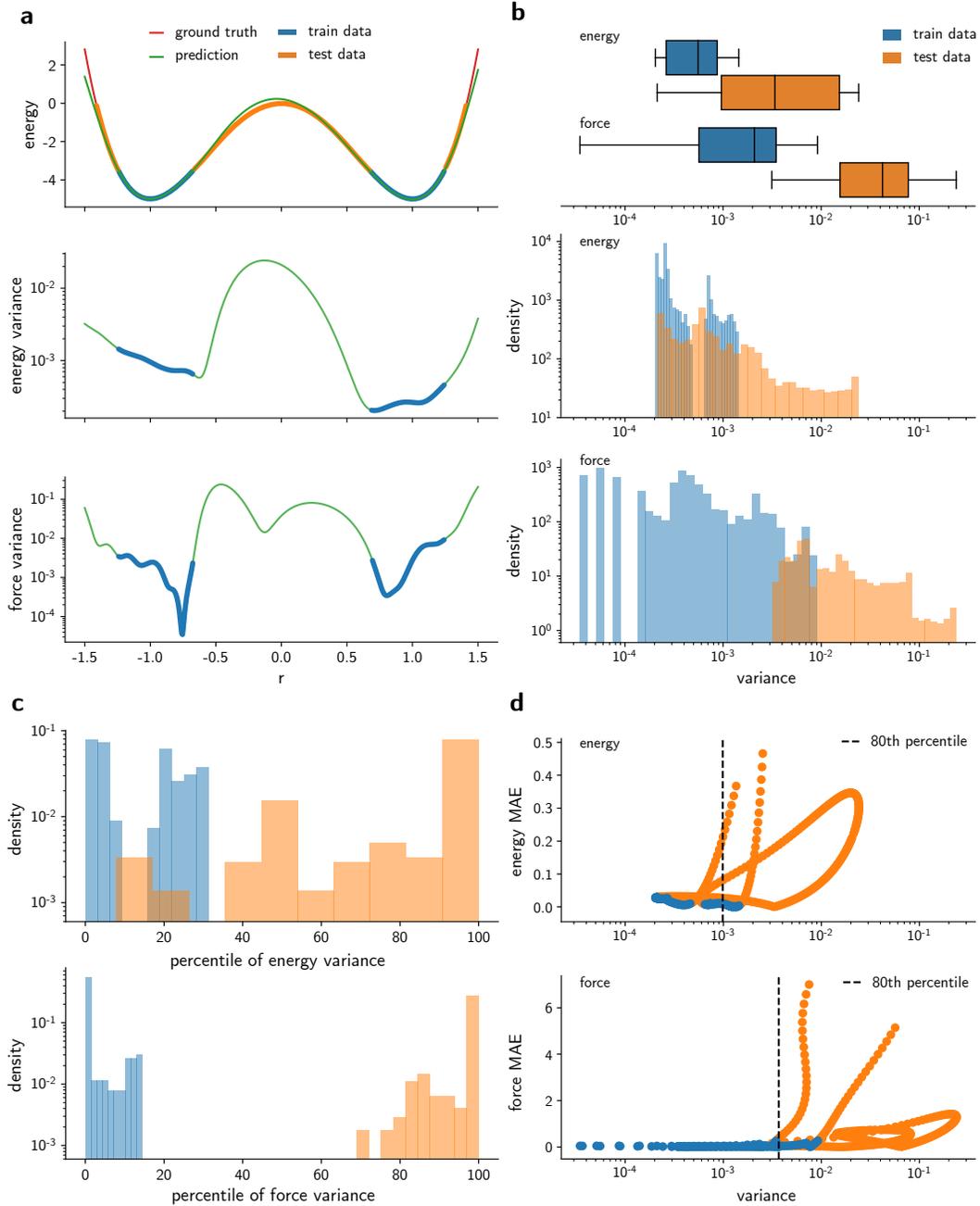

Figure S1. Example of training statistics for the 1D double well potential. **a**, Predictions of energy, energy variance and force variance for five networks trained on the data shown in blue. **b**, Statistics of the models on the train/test data shown in **a**. The vertical line is the median, boxes are the interquartile region, and whiskers represent the range of the distribution. **c**, Distribution of energy and force variances. A smaller overlap between variances of test and train data is found for forces, as opposed to energy predictions. **d**, Relationship between the mean absolute error (MAE) of energies and forces and their corresponding variances. MAE is computed with respect to the ground truth values.

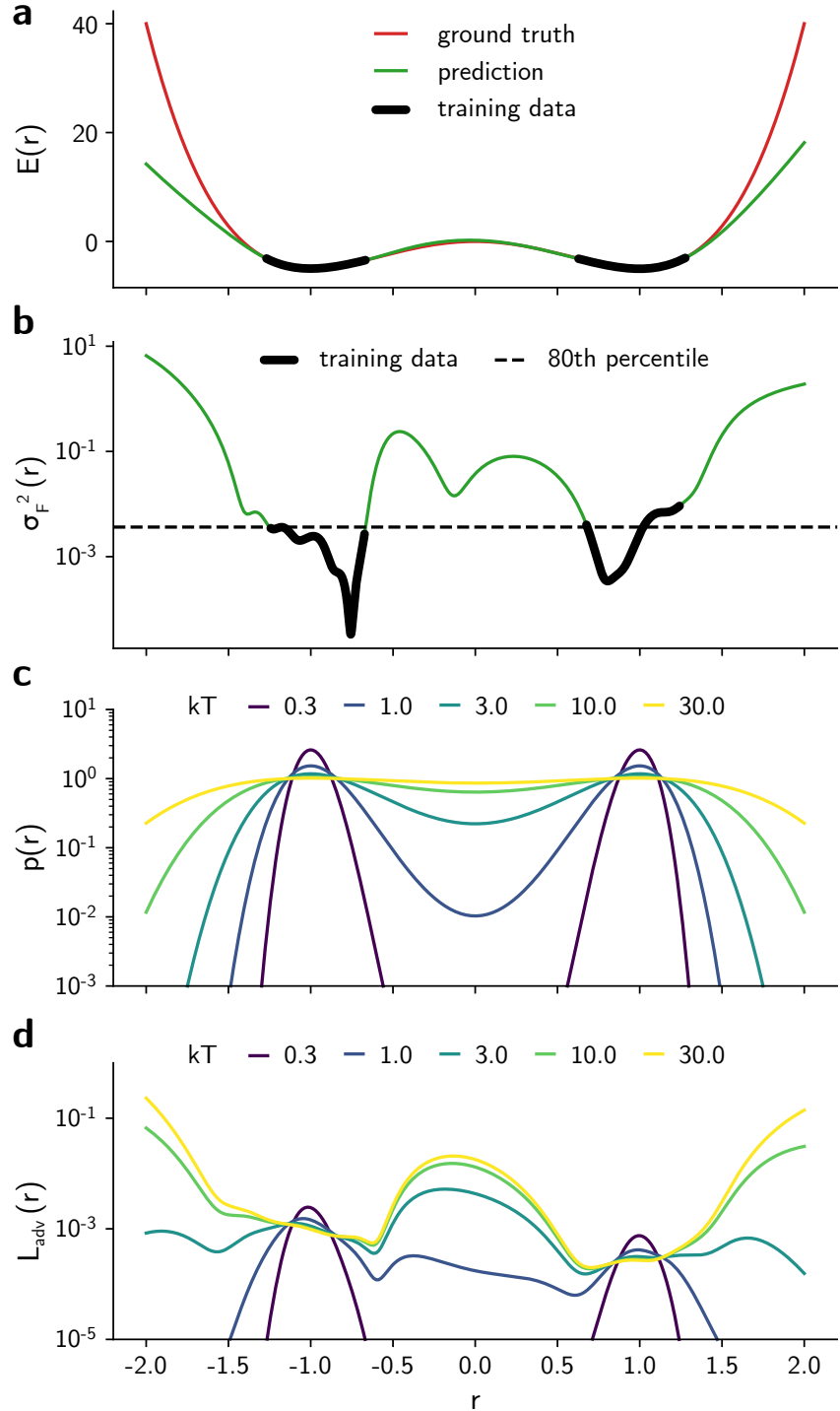

Figure S2. Example of loss function for the example shown in Fig. S1. Predictions of **a**, mean energy and **b**, force variance for the given range of positions $r$. **c**, Boltzmann probability constructed for the ground truth potential. A higher value of $kT$ allows higher energy states to be sampled with higher probability. **d**, Adversarial loss constructed for the neural network potential. For $kT > 3$, the adversarial attack rewards sampling the transition state of the double well potential.

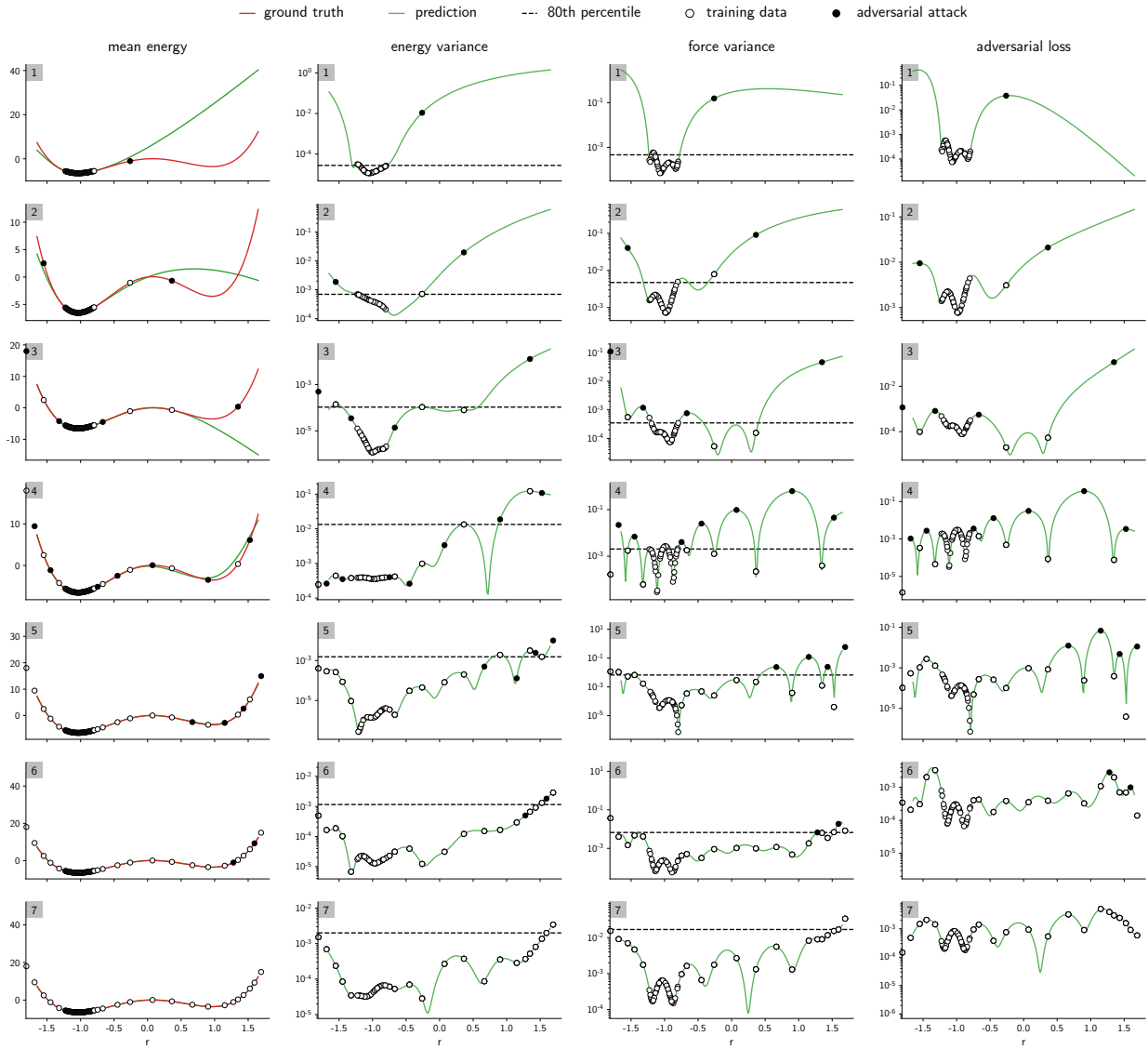

Figure S3. Evolution of mean energy, energy and force variance, and adversarial loss for a one dimensional double well potential. The number of the generation is shown on the top left corner of each plot. The normalized temperature for the adversarial loss is 5. The dashed line is the 80th percentile of the variance of the training data, and is often useful for deduplication.

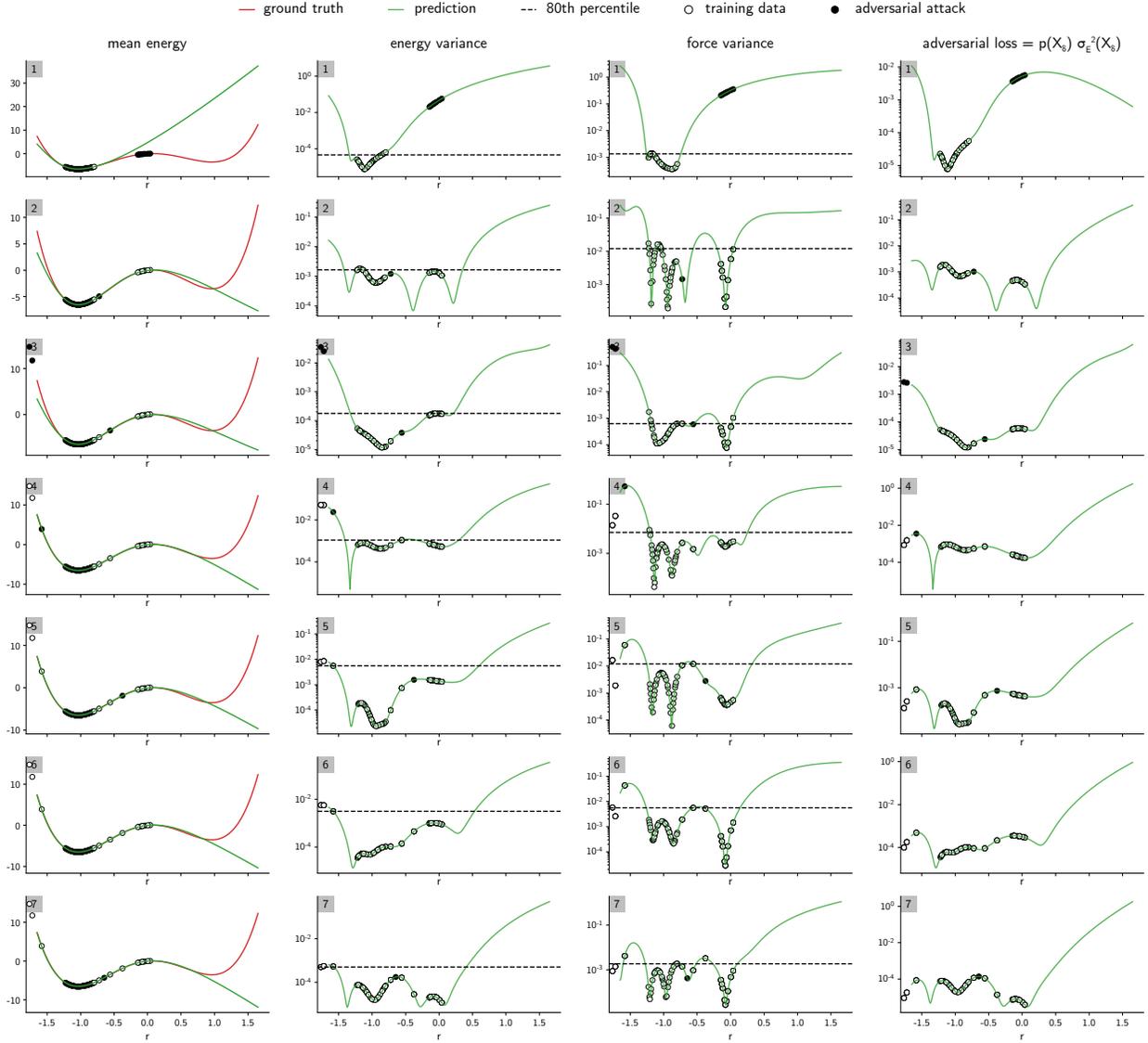

Figure S4. Evolution of mean energy, energy and force variance, and modified adversarial loss for a one dimensional double well potential. The adversarial loss from Eq. (13) of the main paper was modified to use $\sigma_E^2$ instead of $\sigma_F^2$. Due to the force matching approach, the energy uncertainty does not allow the space to be adequately explored. The number of the generation is shown on the top left corner of each plot. The normalized temperature for the adversarial loss is 5. The dashed line is the 80th percentile of the variance of the training data.

S9

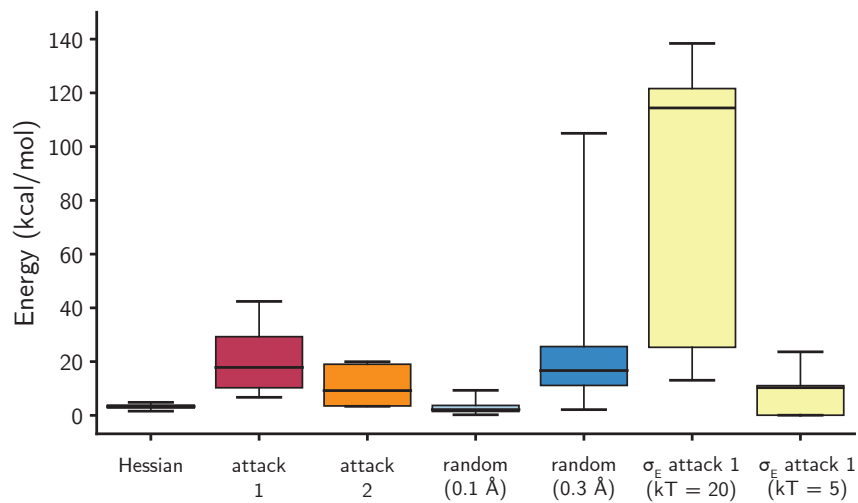

Figure S5. Distribution of DFT energies for conformations of an ammonia molecule sampled with different methods, including adversarial attacks performed using the energy uncertainty ($\sigma_E^2$). The horizontal line is the median, the box is the interquartile region and the whiskers span the range of the distribution.

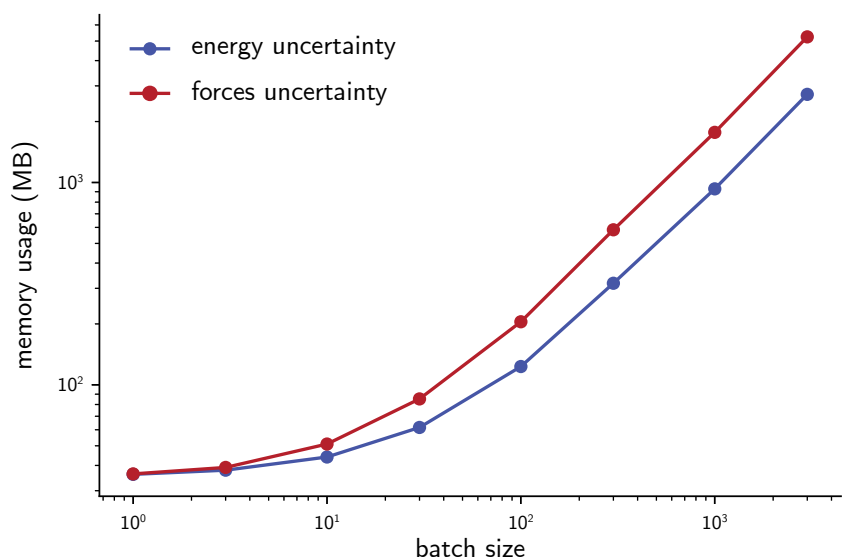

Figure S6. Total GPU memory used by 5 SchNet models, a batch of ammonia molecules, and its computational graph when the uncertainty in energy (blue) and forces (red) are calculated. The cached GPU memory is not taken into consideration in this plot.

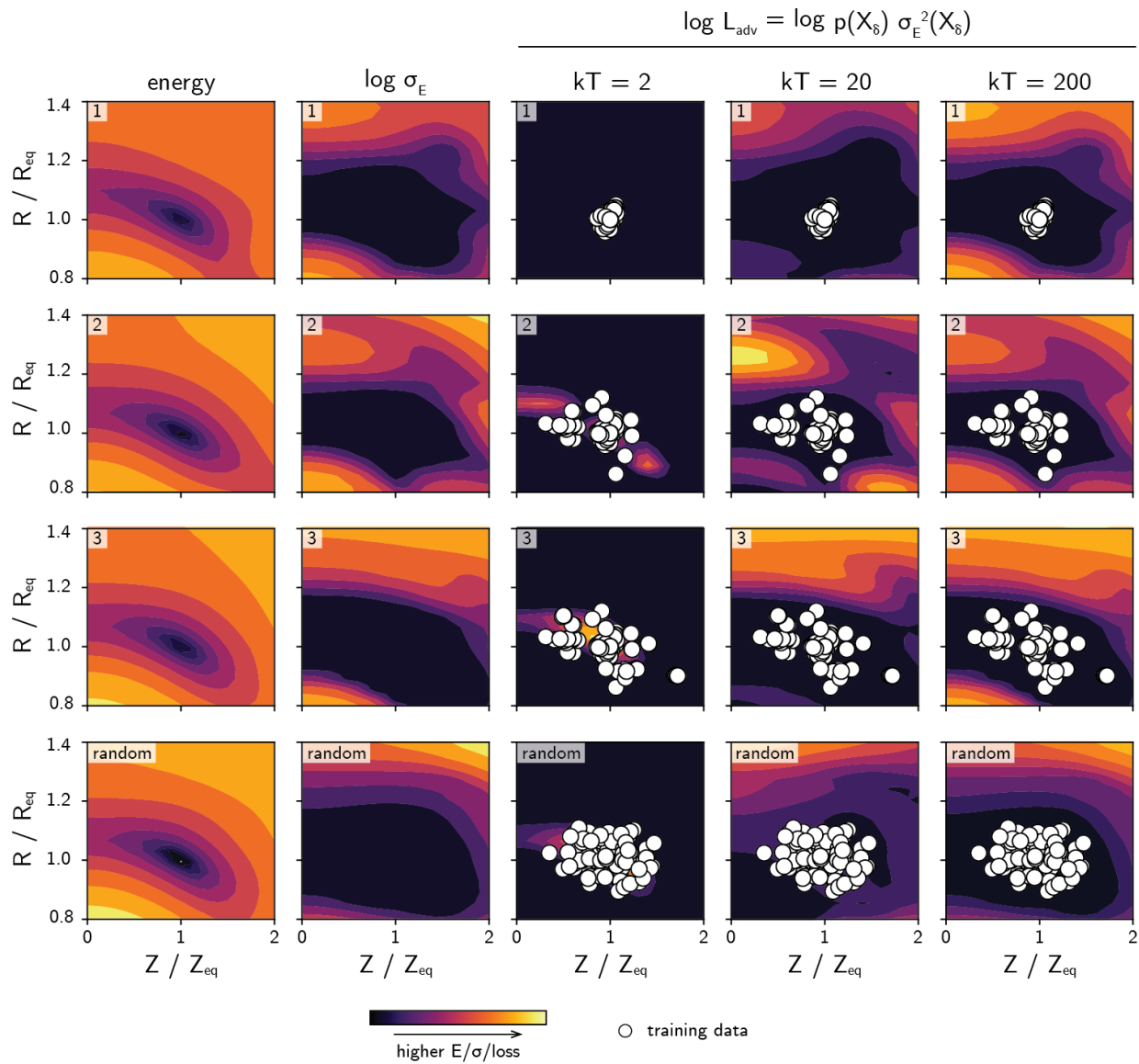

Figure S7. Relationship between the projected PES, energy uncertainty and adversarial loss for ammonia. When the energy variance is used for the adversarial loss function, smaller contrasts between regions within and outside of the training set typically lead to a worse exploration of the phase space when compared to an adversarial loss based on forces uncertainty (see Fig. S8. The number of the generation is shown on the top left corner of each plot. The variables $R$ and $Z$ are defined in Fig. 3 of the main text. $kT$ is given in kcal/mol.

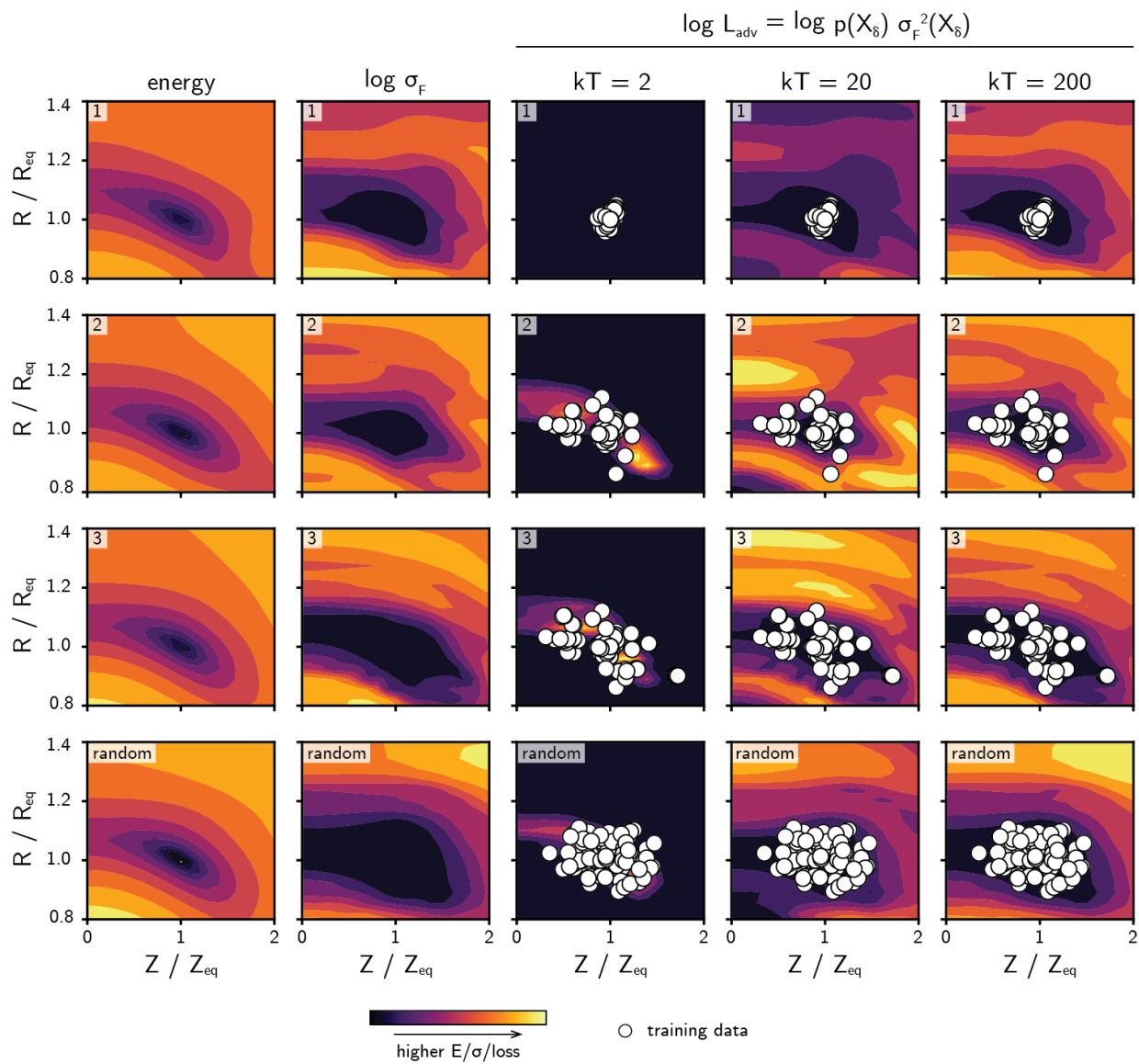

Figure S8. Relationship between the projected PES, energy uncertainty and adversarial loss for ammonia. When the force variance is used for the adversarial loss function, a higher contrast between regions within and outside of the training set favor an informed exploration of the phase space when compared to an adversarial loss based on energy uncertainty (see Fig. S7. The number of the generation is shown on the top left corner of each plot. The variables $R$ and $Z$ are defined in Fig. 3 of the main text. $kT$ is given in kcal/mol.

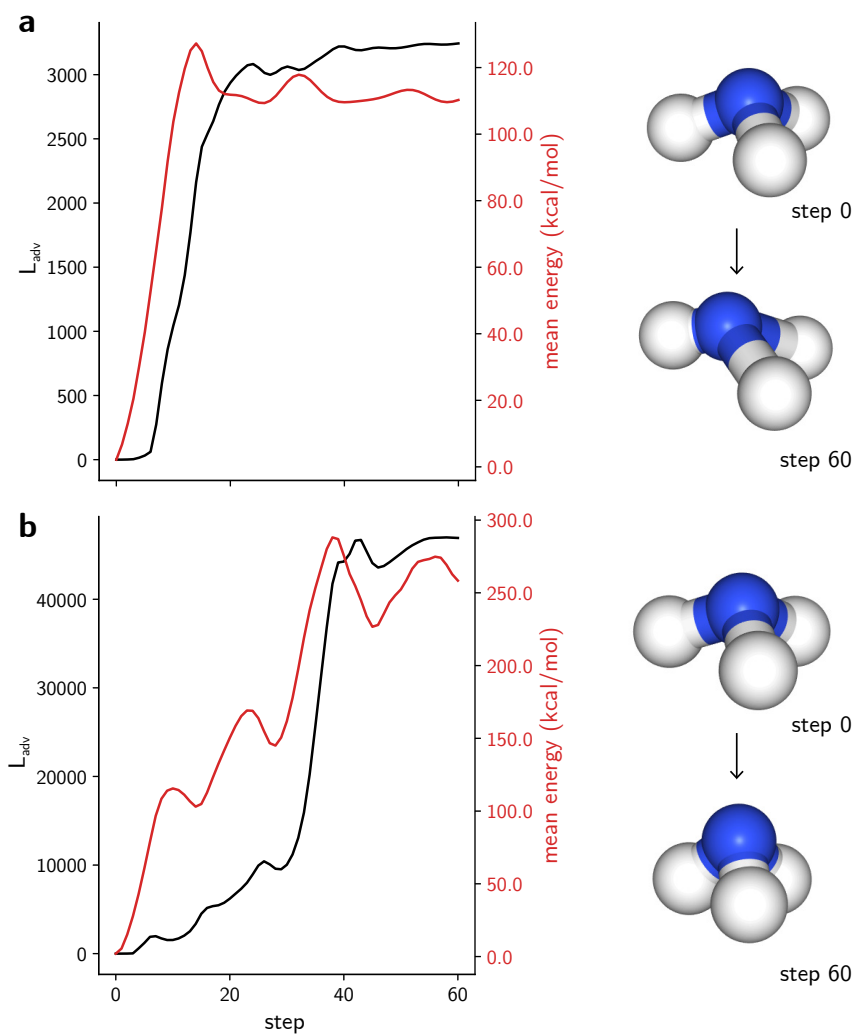

Figure S9. Examples of an adversarial attack on the ammonia molecule using the third generation of NN potentials for this system. Within 60 optimization steps, the adversarial loss converges to higher energy geometries that include **a**, off-center distortions and **b**, symmetrically approaching the hydrogen atoms. The high energies are allowed by the adversarial loss due to a high sampling temperature ($kT = 20$ kcal/mol).

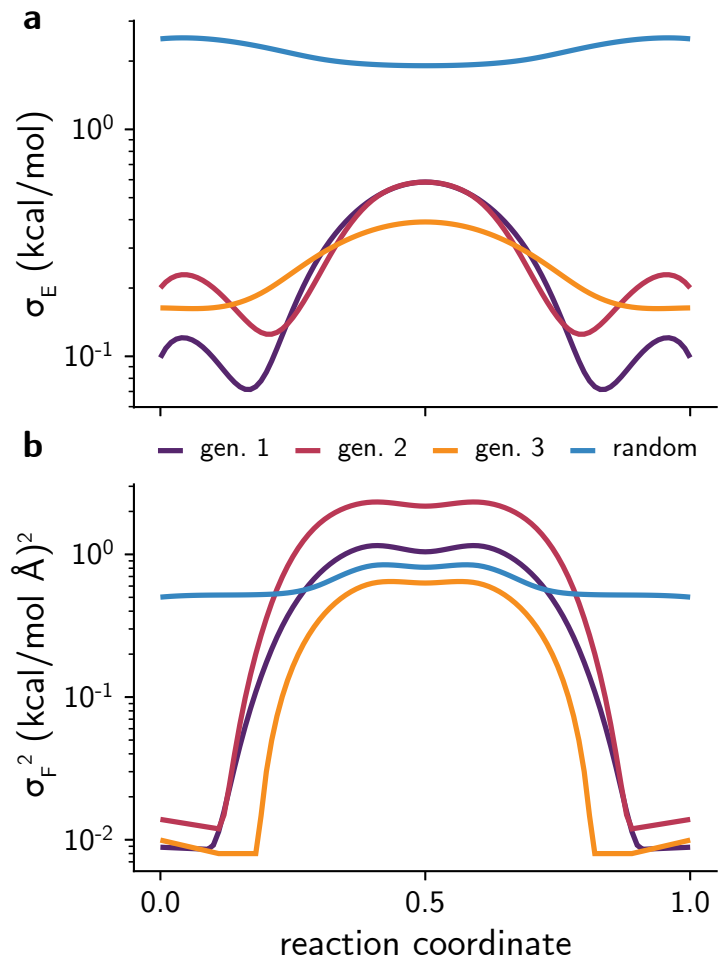

Figure S10. Uncertainty in **a**, energy and **b**, forces for the nitrogen inversion barrier. Whereas the uncertainty close to the barrier tends to lower as the active learning loop progresses, it stays approximately constant for the NN potential trained on randomly-distorted geometries.

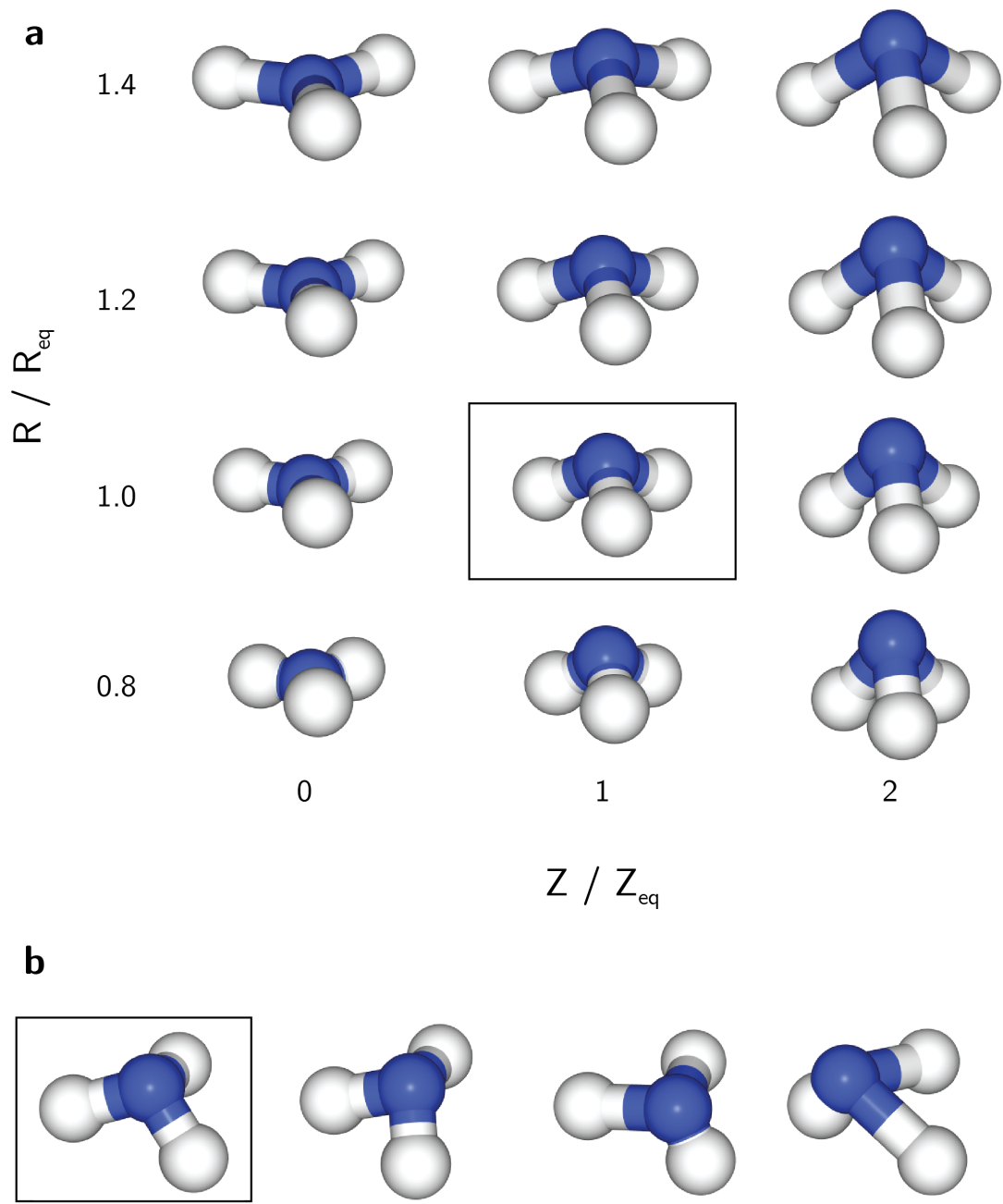

Figure S11. **a**, Example of ammonia geometries created for different values of $(Z, R)$. **b**, Different distortions of the ground state geometry (outlined in black) that have the same values of $(Z_{\text{eq}}, R_{\text{eq}})$

S15

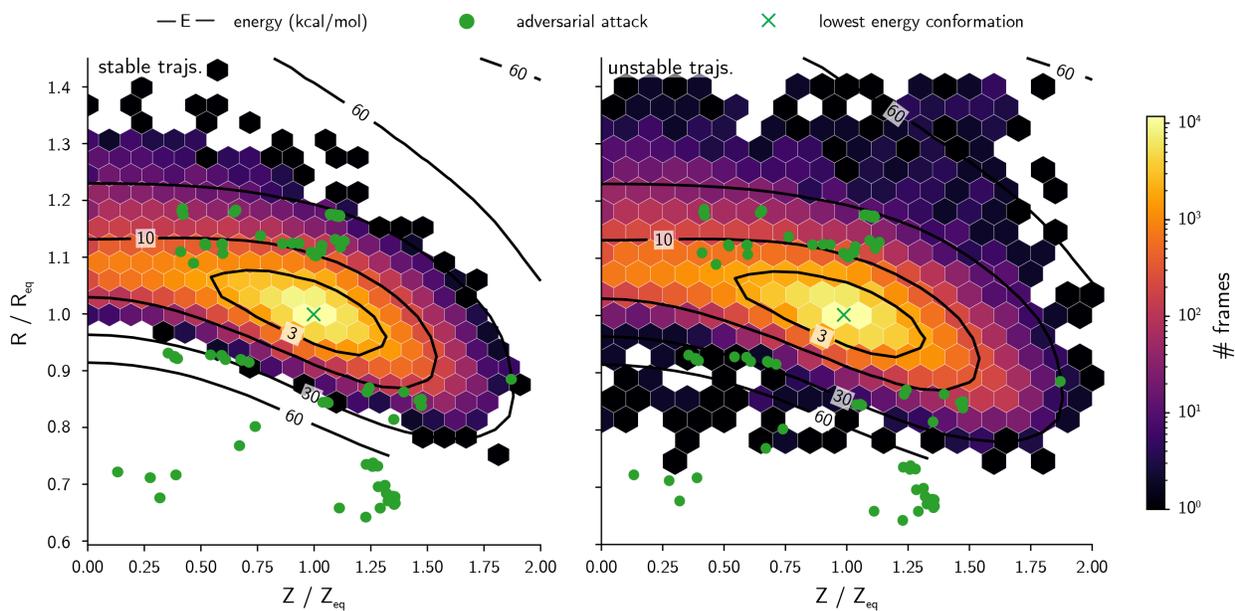

Figure S12. Density of frames for stable (left) and unstable (right) molecular dynamics trajectories obtained with the third generation of NN potentials for ammonia. Contour lines indicate constant energy levels in the phase space. Points for 100 adversarial attacks performed for the third generation indicate that the adversarial strategy samples points not well explored by the NN potential in MD trajectories. Only a subset of the phase space of unstable trajectories is shown, as several data points fall outside of this region.

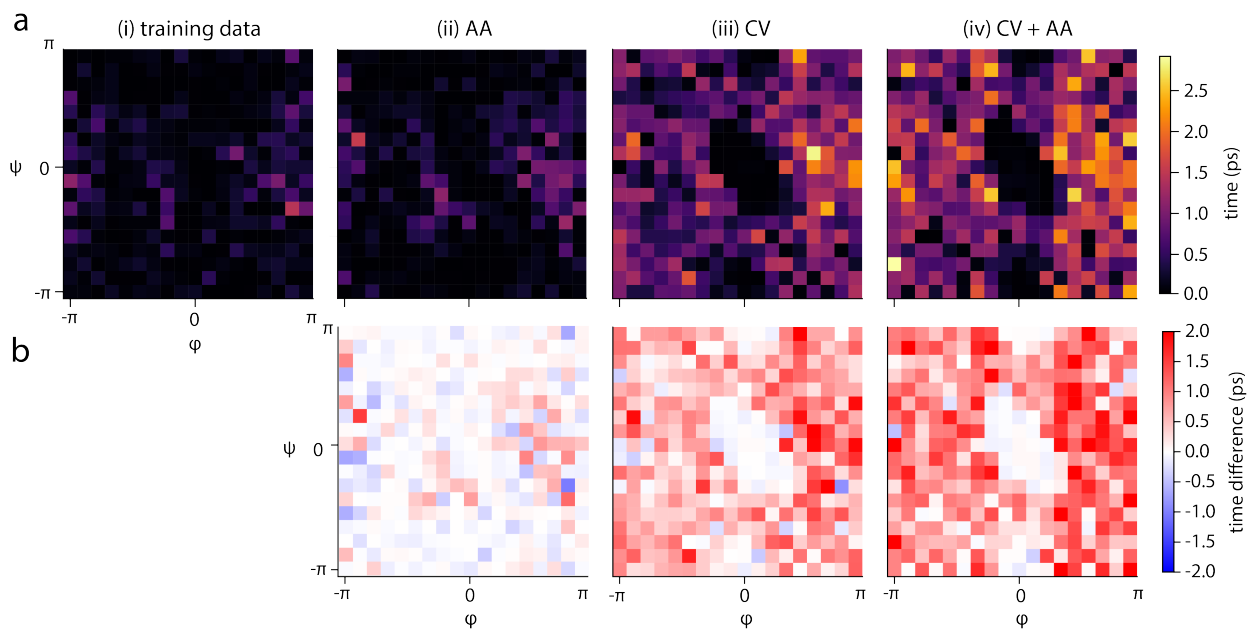

Figure S13. **a**, Duration of stable molecular dynamics trajectories of alanine dipeptide, where grid points correspond to dihedral angles $(\varphi, \psi)$ of starting configurations. MD trajectories are obtained with NN potentials trained on (i) only 10,000 MD data points, (ii) MD data + 3 generations of all-atom (AA) adversarial samples, (iii) MD data + 7 generations of predefined collective variable (CV) attacks, and (iv) MD data + 7 generations of CV attacks followed by 3 generations of AA attacks. **b**, Difference in durations of stable trajectories with respect to (i). Adding CV and AA attacks increases the robustness of NN potentials, allowing longer trajectories to be sampled.

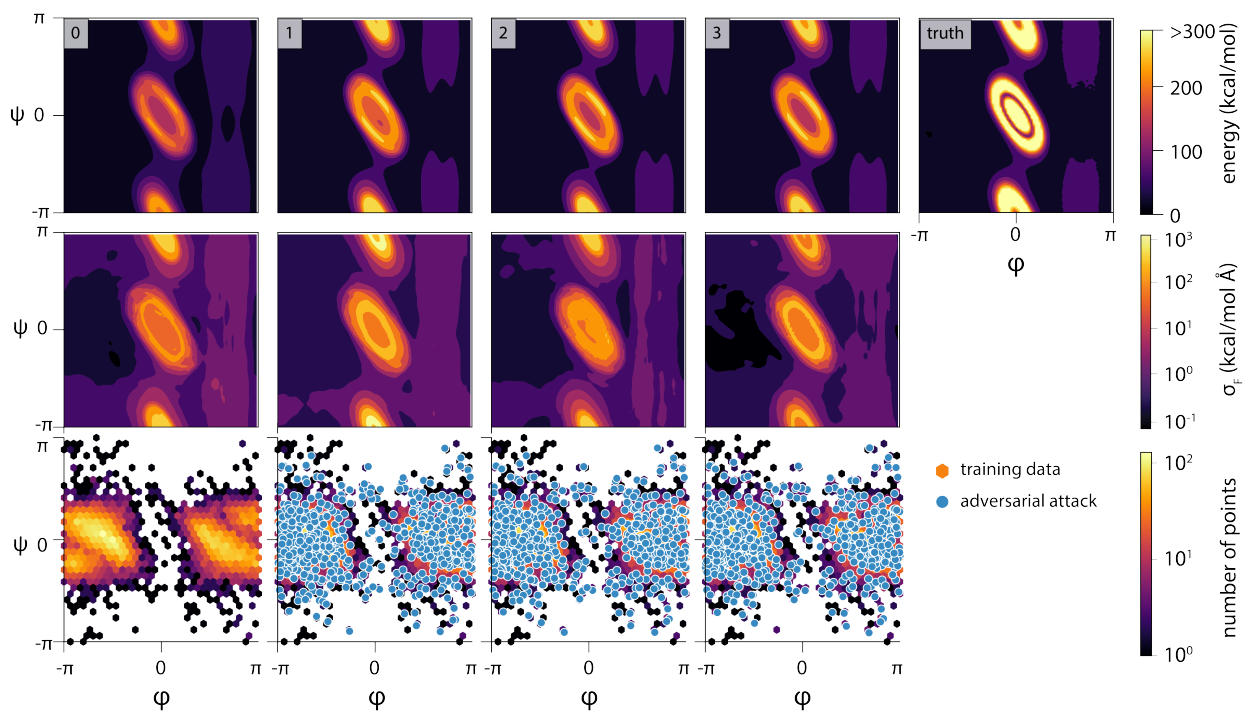

Figure S14. Evolution of PES of an NN committee trained on 3 generations of adversarial examples. Adversarial examples (red points) are sampled by applying small displacements $\delta$ to positions of all atoms (AA). Generation 0 correspond to the same NN committee from generation 7 in the main paper (Fig. 4), where the committee was already trained on configurations from original MD training data and CV adversarial examples (all included as hexagonal bins in figure). Angles are given in radians.



\end{document}